\documentclass[10pt,twocolumn,letterpaper]{article}

\usepackage{cvpr}      
\usepackage{csquotes}

\definecolor{cvprblue}{rgb}{0.21,0.49,0.74}
\usepackage[pagebackref,breaklinks,colorlinks,allcolors=cvprblue]{hyperref}

\def\paperID{13797} 
\def\confName{CVPR}
\def\confYear{2025}
\def\ourmethod{Proto-RSet}
\def\sim{\text{sim}}

\title{Rashomon Sets for Prototypical-Part Networks: Editing Interpretable Models in Real-Time}

\author{Jon Donnelly\\
Duke University\\
{\tt\small jon.donnelly@duke.edu}
\and
Zhicheng Guo\\
Duke University\\
{\tt\small zhicheng.guo@duke.edu}
\and
Alina Jade Barnett\\
Duke University\\
{\tt\small alina.barnett@duke.edu}
\and
Hayden McTavish\\
Duke University\\
{\tt\small hayden.mctavish@duke.edu}
\and
Chaofan Chen\\
University of Maine\\
{\tt\small chaofan.chen@maine.edu}
\and
Cynthia Rudin\\
Duke University\\
{\tt\small cynthia@cs.duke.edu}
}

\begin{document}
\maketitle

\begin{abstract}
Interpretability is critical for machine learning models in high-stakes settings because 
it allows users to verify the model's reasoning. 
In computer vision, prototypical part models (ProtoPNets) have become the dominant model type to meet this need.
Users can easily identify flaws in ProtoPNets, but fixing problems in a ProtoPNet requires slow, difficult retraining that is not guaranteed to resolve the issue. This problem is called the ``interaction bottleneck.''
We solve the interaction bottleneck for ProtoPNets by simultaneously finding many equally good ProtoPNets (i.e., a draw from a ``Rashomon set''). 
We show that our framework -- called \ourmethod{} -- quickly produces many accurate, diverse ProtoPNets, allowing users to correct problems in real time while maintaining performance guarantees with respect to the training set. 
We demonstrate the utility of this method in two settings: 1) removing synthetic bias introduced to a bird-identification model and 2) debugging a skin cancer identification model. 
This tool empowers non-machine-learning experts, such as clinicians or domain experts, to quickly refine and correct machine learning models \textit{without} repeated retraining by machine learning experts.
\end{abstract}

\section{Introduction}
\begin{figure}[h]
    \centering
    \includegraphics[width=\linewidth]{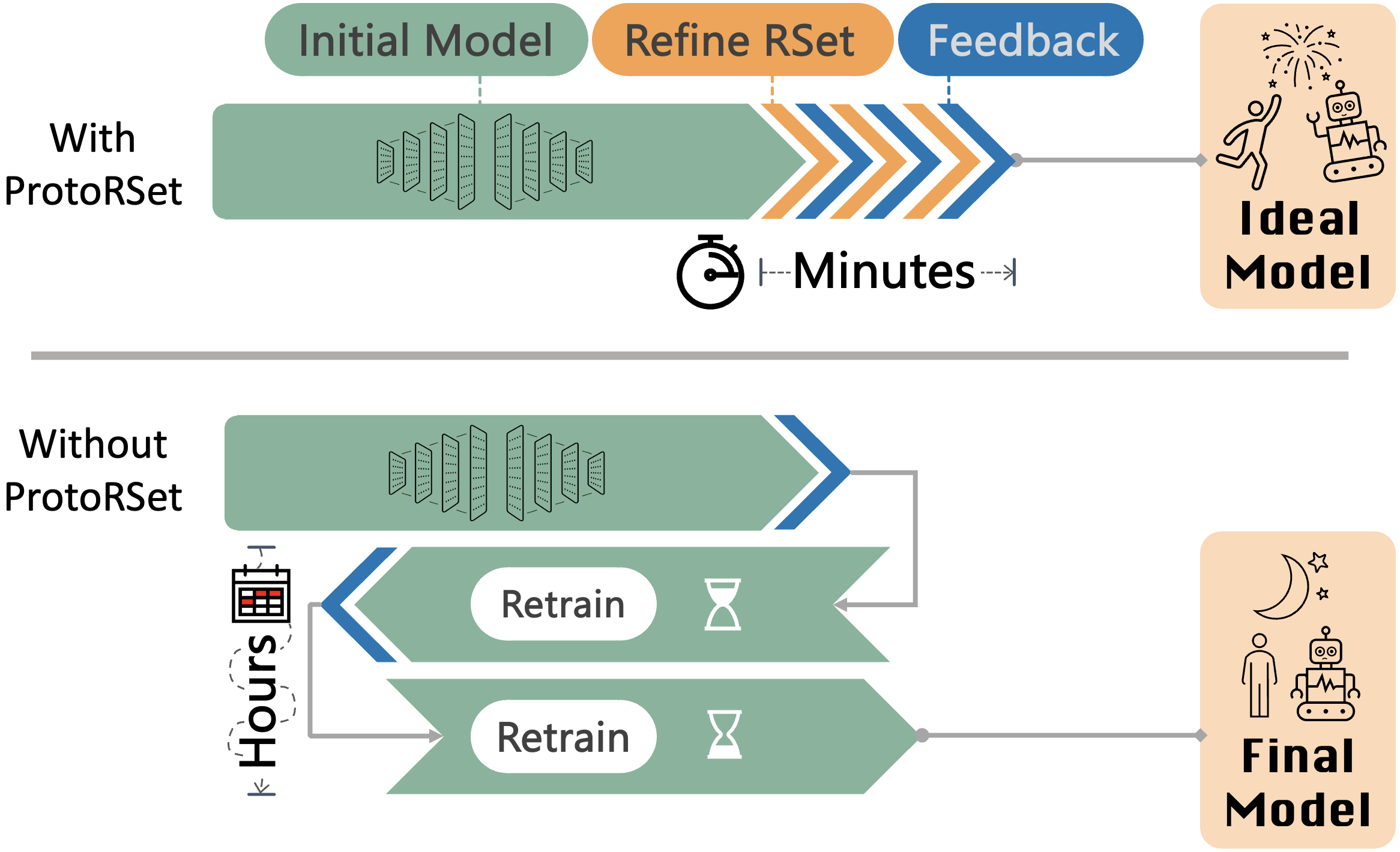}
    \caption{ How \ourmethod{} addresses the interaction bottleneck. (Bottom) Without \ourmethod, incorporating user feedback such as ``this prototype does not make sense, this would be a better option'' requires practitioners to make complicated adjustments to their training regime and train a whole new model. This process can take days, and be prohibitively slow when multiple rounds of feedback are required. (Top) \ourmethod{} allows practitioners to incorporate user feedback \textit{in real time} by selecting different candidate models, eliminating the interaction bottleneck. Moreover, \ourmethod{} guarantees that user constraints are met, producing their ideal model.}
    \label{fig:interaction-bottleneck}
\end{figure}

In high stakes decision domains, there have been increasing calls for interpretable machine learning models \citep{rudin2022interpretable, doshi2017towards, geis2019ethics, com2021laying, us2021artificial}. This demand poses a challenge for image classification, where deep neural networks offer far superior performance to traditional interpretable models. 

Fortunately, case-based deep neural networks have been developed to meet this challenge. These models --- in particular, ProtoPNet \cite{chen2019looks} --- follow a simple reasoning process in which input images are compared to a set of learned prototypes. The similarity of each prototype to a part of the input image is used to form a prediction. This allows users to examine prototypes to check whether the model has learned undesirable concepts and to identify cases that the model considers similar, but that may not be meaningfully related. As such, users can easily identify problems in a trained ProtoPNet.
However, they cannot easily \textit{fix} problems in a ProtoPNet because incorporating feedback into a ProtoPNet requires changing model parameters. In the classic paradigm for model troubleshooting, one would retrain the model with added constraints or loss terms to encode user preferences, but reformulating the problem is time-consuming, and running the algorithm is time-consuming, so repeating this loop even a few times could take days or weeks. This standard troubleshooting paradigm requires alternating input from both domain experts (to provide feedback) and machine learning practitioners (to update the model), which slows down the process. The challenge of troubleshooting models in this classic paradigm has been called the ``interaction bottleneck,'' and it severely limits the ability of domain experts to create desirable models \cite{rudin2024amazing}.

In this work, we solve the interaction bottleneck for ProtoPNets using a different paradigm. We pre-compute many equally-performant ProtoPNets and allow the domain expert to interact with those. Machine learning experts only need to compute the initial set of models, after which the domain expert can interact with the generated models without interruption. 
To make this problem computationally feasible, we compute a set of near-optimal models that use \textit{positive reasoning} over a ProtoPNet with a fixed backbone and set of prototypes. 
Human interaction with this set is easy and near-instantaneous, allowing the user to choose models that agree with their domain expertise and debug ProtoPNets easily. 
The set of near-optimal models for a given problem is called the \textit{Rashomon set}, and thus our approach is called \ourmethod. While recent work has estimated Rashomon sets for tabular data problems \cite{xin2022exploring,zhong2024exploring}, this is the first time it has been attempted for computer vision. 
Thus, it is the first time users are able to interact nimbly with such a large set of complex models.
Figure \ref{fig:interaction-bottleneck} illustrates the classic model troubleshooting paradigm (lower subfigure), as well as the troubleshooting paradigm from \ourmethod{} (upper subfigure), in which we simply provide models that meet a user's constraints from this set rather than retraining an entire ProtoPNet; the calculation takes seconds, not days. 

Our method for building many good ProtoPNets is tractable, as are our methods to filter and sample models from this set based on user preferences. We show experimentally that ProtoRSet is feasible to compute in terms of both memory and runtime, and that ProtoRSet allows us to quickly generate many accurate models that are guaranteed to meet user constraints. Finally, we provide two case studies demonstrating the real world utility of \ourmethod: a user study in which users apply \ourmethod{} to rapidly correct biases in a model, and a case study in which we use \ourmethod{} to simplify a skin cancer classification model.

\section{Related Work}
\subsection{Interpretable Image Classification}
In recent years, a variety of inherently interpretable neural networks for image classification have been developed \cite{bohle2022b, koh2020concept, you2023sum, taesiri2022visual}, but case-based interpretable models \cite{li2018deep, chen2019looks} have become particularly popular. In particular, ProtoPNet introduced a popular framework in which images are classified by comparing parts of the image to learned prototypical parts associated with each class.
A wide array of extensions have followed the original ProtoPNet \cite{chen2019looks}. The majority of these works focus on improving components of the ProtoPNet itself \cite{donnelly2022deformable, nauta2021neural, rymarczyk2022interpretable, rymarczyk2021protopshare, wang2023learning, ma2024looks, wang2021interpretable, nauta2021looks}, improving the training regime \cite{rymarczyk2023icicle, nauta2023pip, willard2024looks},  or applying ProtoPNets to high stakes tasks \cite{yang2024fpn, barnett2021case, choukali2024pseudo, wei2024mprotonet, barnett2024improving}. In principle, \ourmethod{} can be combined with features from many of these extensions, particularly those that use a linear layer to map from prototype activations to class predictions.

Recently, a line of work integrating human feedback into ProtoPNets has developed. ProtoPDebug \cite{bontempelli2023concept} introduced a framework to allow users to inspect a ProtoPNet to determine if changes might help, then to remove or to require prototypes by re-training the network with loss terms aiming to meet these constraints. The R3 framework \cite{liimproving} used human feedback to develop a reward function, which can be used to form loss terms and guide prototype resampling. In contrast to these approaches, \ourmethod{} \textit{guarantees} that accuracy will be maintained, and the constraints given by a user will be met whenever it is possible for such constraints to be met with a well performing model. Our method does not require retraining a neural network.

\subsection{The Rashomon Effect}
In this work, we leverage the \textit{Rashomon effect} to find many near-optimal ProtoPNets. The Rashomon effect refers to the observation that, for a given task, there tend to be many disparate, equally good models \cite{breiman2001statistical}. This phenomenon presents both challenges and opportunities: the Rashomon effect leads to the related phenomenon of predictive multiplicity \cite{marx2020predictive, watson2023multi, kulynych2023arbitrary, hsu2022rashomon, watson2023predictive}, wherein equally good models may yield different predictions for any individual, but has also lead to theoretical insights into model simplicity \cite{semenova2024path, semenova2022existence, boner2024using} and been applied to produce robust measures of variable importance \cite{donnelly2023rashomon, smith2020model, fisher2019all, dong2019variable}. A more thorough discussion of the implications of the Rashomon effect can be found in \cite{rudin2024amazing}.

The set of near-optimal models is the \textit{Rashomon set}. The Rashomon set is defined with respect to a specific model class \cite{semenova2022existence}. In recent years, algorithms have been developed to compute or estimate the Rashomon set over decision trees \cite{xin2022exploring}, generalized additive models \cite{zhong2024exploring}, and risk scores \cite{liu2022fasterrisk}. These methods solve deeply challenging computational tasks, but all concern binary classification for tabular data. In this work, we introduce the first method to approximate the Rashomon set for computer vision problems, though our work also applies to other types of signals that are typically analyzed using convolutional neural networks, including, for instance, medical time series (PPG, ECG, EEG).

\section{Methods}
\subsection{Review of ProtoPNets}

Before describing \ourmethod, we first briefly describe ProtoPNets in general, and the training regime followed in computing a reference ProtoPNet.
Let $\mathcal{D}:= \{\mathbf{X}_i, y_i\}_{i=1}^n$, where $\mathbf{X}_i \in \mathbb{R}^{c \times h \times w}$ is an input image and $y_i \in \{0, 1, \hdots, t - 1\}$ is the corresponding label. Here, $n$ denotes the number of samples in the dataset, $c$ the number of input channels in each image, $h$ the input image height, $w$ the input image width, and $t$ the number of classes.

A ProtoPNet is an interpretable neural network consisting of three components: 
\begin{itemize}
    \item A backbone $f: \mathbb{R}^{c \times h \times w} \to \mathbb{R}^{c' \times h' \times w'}$ that extracts a latent representation of each image
    \item A prototype layer $g: \mathbb{R}^{c' \times h' \times w'} \to \mathbb{R}^{m}$ that computes the similarity between each of $m$ prototypes $\mathbf{p}_j \in \mathbb{R}^{c'}$ and the latent representation of a given image, i.e. $g_j(f(\mathbf{X}_i)) = \max_{a, b} \sim(\mathbf{p}_j, f(\mathbf{X}_i)_{:, a, b}))$ for some similarity metric $\sim,$ where $a, b$ are coordinates along the height and width dimension.\footnote{In practice, prototypes may consist of multiple spatial components and therefore compare to multiple locations at a time, but for simplicity we only consider prototypes that have spatial size $(1 \times 1)$ in this paper.}
    \item A fully connected layer $h: \mathbb{R}^{m} \to \mathbb{R}^{t}$ that computes an overall classification based on the given prototype similarities. Here, $h$ outputs valid class probabilities (i.e., contains a softmax over class logits).
\end{itemize} 

These layers are optimized for cross entropy and several other loss terms (see \cite{chen2019looks} for details) using stochastic gradient descent. At inference times, each predicted class probability vector is formed as the composition $\hat{\mathbf{y}}_i = h \circ g \circ f(\mathbf{X}_i)$.

\subsection{Estimating the Rashomon Set}
In general, the Rashomon set of ProtoPNets is defined as:
\begin{align}
\label{def:ideal_rset}
\begin{split}
    \mathcal{R}(&\mathcal{D}_{train}; \theta, \ell, \lambda)\\ 
    :=\{&(\mathbf{w}_f, \mathbf{w}_g, \mathbf{w}_h)\\
    &: \ell\left(h\left(g(f(\cdot; \mathbf{w}_f); \mathbf{w}_g);\mathbf{w}_h\right), \mathcal{D}_{train}; \lambda\right) \leq \theta\},
\end{split}
\end{align}
where $\ell$ is any regularized loss, $\lambda$ is the weight of the regularization, $\theta$ is the maximum loss allowed in the Rashomon set, and each term $\mathbf{w}$ denotes all parameters associated with the subscripted layer. While $\mathcal{R}$ would be of great practical use, it is intractable to compute: $f$ is typically an extremely complicated, non-convex function, and $\mathbf{w}_f$ and $\mathbf{w}_g$ are extremely high dimensional.

However, if we fix $\mathbf{w}_f$ and $\mathbf{w}_g$ to some reasonable reference values $\bar{\mathbf{w}}_f$ and $\bar{\mathbf{w}}_g$ and instead target
\begin{align}
\label{def:rset_wrt_lastlayer}
\begin{split}
    &\bar{\mathcal{R}}(\mathcal{D}_{train}; \theta, \ell, \lambda) :=\\
    &\{\mathbf{w}_h: \ell\left(h\left(g(f(\cdot; \bar{\mathbf{w}}_f); \bar{\mathbf{w}}_g);\mathbf{w}_h\right), \mathcal{D}_{train}; \lambda\right) \leq \theta\},
\end{split}
\end{align}
we arrive at a much more approachable problem that supports many of the same use cases as $\mathcal{R}$. In particular, we have reduced (\ref{def:ideal_rset}) to the problem of finding the Rashomon set of multiclass logistic regression models. We describe methods to compute these reference values, with use-cases in the following section, but first let us introduce a method to approximate $\bar{\mathcal{R}}$.

For simplicity of notation, let $\bar{\ell}(\mathbf{w}_{h}) := \ell\left(h\left(g(f(\cdot; \bar{\mathbf{w}}_{f}); \bar{\mathbf{w}}_{g});\mathbf{w}_{h}\right), \mathcal{D}_{train}; \lambda\right);$ that is, the loss of a model where $h$ is parametrized by $\mathbf{w}_h$ and all other components of the ProtoPNet are fixed. 

Note that optimizing $\bar{\ell}(\mathbf{w}_h)$ for $\mathbf{w}_h$ is exactly optimizing multiclass logistic regression, which is a convex problem. As such, we can compute the optimal coefficient value $\mathbf{w}_h^*$ with respect to $\bar{\ell}$ using gradient descent.
Inspired by \cite{zhong2024exploring}, we approximate the loss for any coefficient vector $\mathbf{w}_h$ using a second order Taylor expansion:
\begin{align*}
    \bar{\ell}(&\mathbf{w}_h) \\\approx &
    \bar{\ell}(\mathbf{w}_h^*) + ({\nabla \bar{\ell}|_{\mathbf{w}_h^*}})^{T} (\mathbf{w}_h - \mathbf{w}_h^*)\\ &+ \frac{1}{2}(\mathbf{w}_h - \mathbf{w}_h^*)^T\mathbf{H}(\mathbf{w}_h - \mathbf{w}_h^*)\\
    =&\bar{\ell}(\mathbf{w}_h^*) + \frac{1}{2}(\mathbf{w}_h - \mathbf{w}_h^*)^T\mathbf{H}(\mathbf{w}_h - \mathbf{w}_h^*),
\end{align*}
where $\mathbf{H}$ denotes the Hessian of $\bar{\ell}$ with respect to $\mathbf{w}_h$. The above equality holds because $\mathbf{w}_h^*$ is the loss minimizer, making $\nabla \bar{\ell}|_{\mathbf{w}_h^*} = \mathbf{0}.$ 
Plugging the above formula for $\bar{\ell}(\mathbf{w}_h)$ into (\ref{def:rset_wrt_lastlayer}), we are interested in finding each $\mathbf{w}_h$ such that
\begin{align*}
    \frac{1}{2(\theta - \bar{\ell}(\mathbf{w}_h^*))}(\mathbf{w}_h - \mathbf{w}_h^*)^T\mathbf{H}(\mathbf{w}_h - \mathbf{w}_h^*) \leq 1;
\end{align*}
this set is an ellipsoid with center $\mathbf{w}_h^*$ and shape matrix $\frac{1}{2(\theta - \ell(\mathbf{w}_h^*))} \mathbf{H}.$ 

This object is convenient to interact with (see Subsection \ref{subsec:interaction}), but it poses computational challenges. For a standard fully connected layer $h$, we have $\mathbf{w}_h \in \mathbb{R}^{mt}$ and $\mathbf{H} \in \mathbb{R}^{mt \times mt}$ -- this is a prohibitively large matrix, which requires $\mathcal{O}(m^2t^2)$ floats in memory. The original ProtoPNet \cite{chen2019looks} used 10 prototypes per class for $200$ way classification; with this configuration, a Hessian consisting of 32 bit floats would require 5.12 terabytes to store. In Appendix \ref{supp:sampling_no_hessian}, we describe how models may be sampled from this set using $\mathcal{O}(m^2 + t^2)$ floats in memory.

We can reduce these computational challenges while adhering to a common practitioner preference by requiring \textit{positive reasoning} (i.e., this is class $a$ because it looks like a prototype from class $a$) rather than \textit{negative reasoning} (i.e., this is class $a$ because it does not look like a prototype from class $b$). We consider the Rashomon set defined over parameter vectors $\mathbf{w}_h$ that \textit{only allow positive reasoning}, restricting all connections between prototypes and classes other than their assigned class to $0$. This allows us to store a Hessian $\mathbf{H} \in \mathbb{R}^{m \times m},$ substantially reducing the memory load -- in the case above, from 5.12 terabytes to 128 megabytes. Appendix \ref{supp:pos_only_hessian} provides a derivation of the Hessian for these parameters. 


\subsection{Interacting With the Rashomon Set}
\label{subsec:interaction}
We introduce methods for three common interactions with a \ourmethod: 1) sampling models from a \ourmethod; 2) finding a subset of models that \textit{do not} use a given prototype; and 3) finding a subset of models that uses a given prototype with coefficient at least $\alpha$.

\textbf{Sampling models from a \ourmethod}: First, we select a random direction vector $\mathbf{d} \in \mathbb{R}^m$ and a random scalar $\kappa \in [0, 1].$ We will produce a parameter vector $\hat{\mathbf{w}}_h := \tau \mathbf{d} + \mathbf{w}_h^*$, where $\tau$ is the value such that:
\begin{align*}
    \frac{1}{2(\theta - \ell(\mathbf{w}_h^*))} \tau^2\mathbf{d}^T \mathbf{H} \mathbf{d} = \kappa\\
    \iff \tau = \sqrt{\frac{2\kappa(\theta - \ell(\mathbf{w}^*_h))}{\mathbf{d}^T \mathbf{H} \mathbf{d}}}
\end{align*}
The resulting vector $\hat{\mathbf{w}}_h$ is the result of walking $\kappa$ proportion of the distance from $\mathbf{w}_h^*$ to the border of the Rashomon set along direction $\mathbf{d}$. This operation allows users to explore individual, equally viable models and get a sense for what a given \ourmethod \phantom{} will support.

\textbf{Finding the subset of a \ourmethod{} that does not use a given prototype}: We say a ProtoPNet does not use a prototype if every weight assigned to that prototype in the last layer is $0$. We leverage the fact that \ourmethod{} is an ellipsoid to reframe this as the problem of finding the intersection between a hyperplane and an ellipsoid. In particular, to remove prototype $j,$ we define a hyperplane
\begin{align*}
    \mathcal{H}_j := \{\mathbf{w}_h : \mathbf{e}_j^T\mathbf{w}_h = 0\},
\end{align*}
where $\mathbf{e}_j$ is a vector of zeroes, save for a $1$ at index $j.$ Thus, to remove prototype $j$, we compute $\bar{\mathcal{R}}_{reduced} = \mathcal{H}_j \cap \bar{\mathcal{R}}$ using the method described in \cite{kurzhanskiy2006ellipsoidal}. Note that the intersection between a hyperplane and an ellipsoid is still an ellipsoid, meaning we can still easily perform all operations described in this section on  $\bar{\mathcal{R}}_{reduced},$ including removing multiple prototypes. This operation is useful if, for example, a user wants to remove a prototype that encodes some unwanted bias or confounding.

This procedure provides a natural check for whether or not it is viable to remove a prototype. If $\mathcal{H}_j$ and $\bar{\mathcal{R}}$ do not intersect, it means that prototype $\mathbf{p}_j$ cannot be removed while remaining in the Rashomon set. This feedback can then be provided to the user, helping them to understand which prototypes are essential for the given problem.

\begin{figure}
    \centering
    \includegraphics[width=1.0\linewidth]{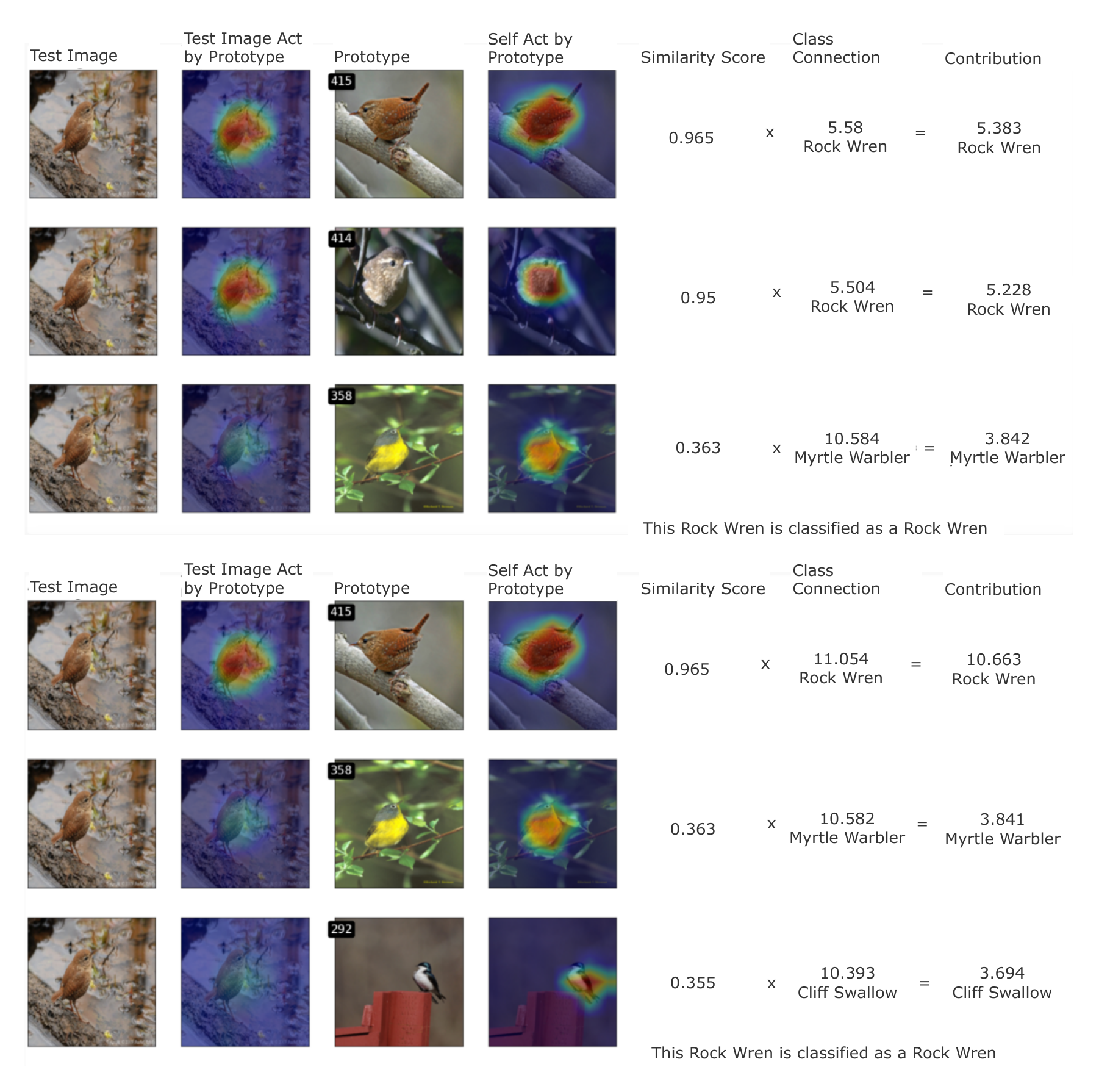}
    \caption{Models produced by ProtoRSet before (top) and after (bottom) a user specifies that prototype 414 must be removed. \ourmethod{} guarantees that the bottom model has similar performance to the top, despite following a different reasoning process. If a prototype cannot be removed while maintaining performance, \ourmethod{} quickly identifies and reports this.}
    \label{fig:removing_proto}
\end{figure}

These removals often result in non-trivial changes to the overall reasoning of the model --- this is not just setting some coefficients to zero. Figure \ref{fig:removing_proto} provides a real example of this phenomena. When prototype 414 is removed, the weight assigned to prototype 415 is roughly doubled to account for this change in the model, despite being a fairly different prototype.

\textbf{Finding the subset of a \ourmethod{} that uses a prototype with coefficient at least $\alpha$}: 
Here, we leverage the fact that \ourmethod{} is an ellipsoid to reframe this to the problem of finding the intersection between a half-space and an ellipsoid. In particular, to force prototype $j$ to have coefficient $w^{(h)}_j \geq \alpha,$ we are interested in the half-space:
\begin{align*}
    \mathcal{S}_j(\alpha) := \{\mathbf{w}_h : \mathbf{e}_j^T\mathbf{w}_h \geq \alpha\}.
\end{align*}
Note that the intersection between an ellipsoid and a half-space is \textbf{not} an ellipsoid, meaning we cannot easily perform the operations described in this section on  $\mathcal{S}_j(\alpha) \cap \bar{\mathcal{R}}.$ As such, we do not apply these constraints until after prototype removal. We describe a regime to sample models from the Rashomon set after applying multiple halfspace constraints by solving a convex quadratic program in Appendix \ref{supp:sampling_halfspace}. This operation is useful if, for example, a user finds a prototype that has captured some critical information according to their domain expertise.

\subsection{Sampling Additional Candidate Prototypes}
\label{subsec:sampling_prototypes}
When too many constraints are applied, there may be no models in the existing Rashomon set that satisfy all criteria. Instead of retraining or restarting the model selection process, we can sample additional prototypes in order to expand the Rashomon set. This way, we can retain all existing feedback and continue model refinement.

We generate $s$ new candidate prototypes by randomly selecting patches from the latent representations of the training set images, as we now describe. The backbone $f: \mathbb{R}^{c \times h \times w} \to \mathbb{R}^{c' \times h' \times w'}$ extracts a latent representation of each image. We randomly select an image $i$ from all training images, and from $f(\mathbf{X}_i)$, we randomly select one $c'$-length vector from the $h' \times w'$ latent representation. This random selection is repeated $s$ times to provide our $s$ prototype vectors. We then recompute our \ourmethod{} with respect to this augmented set of prototypes, and reapply all constraints generated by the user to this point. 


\section{Experiments}
\begin{figure*}[h!]
    \centering
    \includegraphics[width=0.8\linewidth]{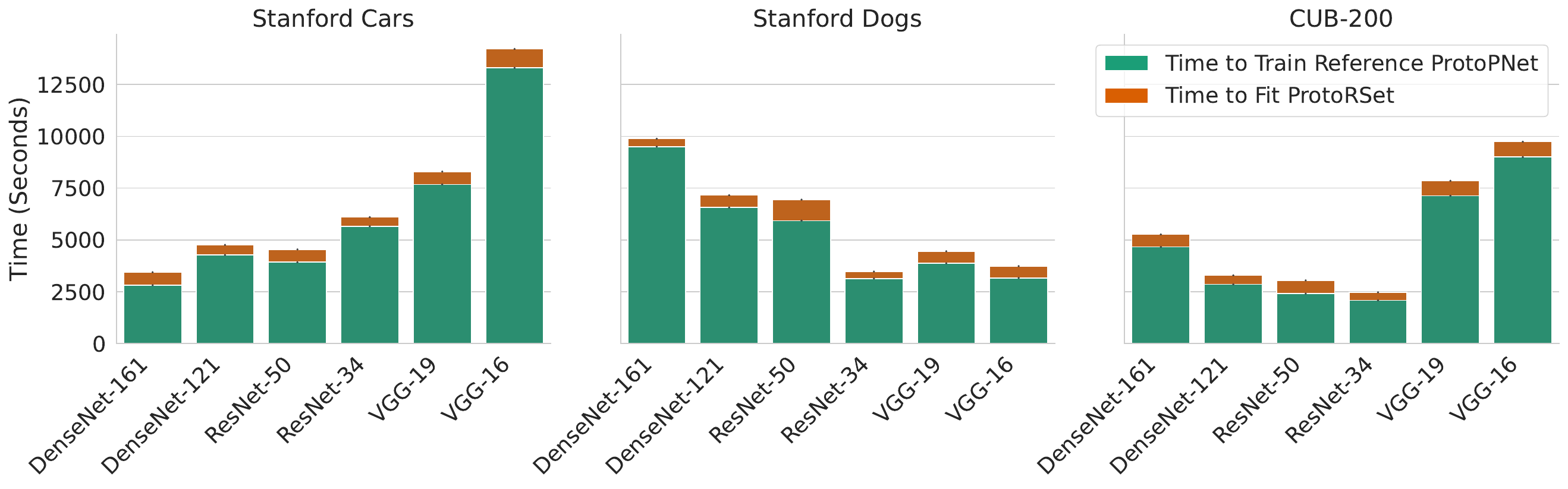}
    \caption{Time to compute \ourmethod{} across three datasets and six backbones as a stacked bar plot. The lower bar presents the time required to train the base ProtoPNet, and the top represents the time to compute \ourmethod{} given that reference ProtoPNet. We find that \ourmethod{} can be computed in less than twenty minutes across a variety of settings.}
    \label{fig:rset-comp-time}
\end{figure*}

We evaluate \ourmethod{} over three fine-grained image classification datasets (CUB-200 \cite{WahCUB_200_2011}, Stanford Cars \cite{KrauseStarkDengFei-Fei_3DRR2013}, and Stanford Dogs \cite{khosla2011novel}), with ProtoPNets trained on six distinct CNN backbones (VGG-16 and VGG-19 \cite{vgg}, ResNet-34 and ResNet-50 \cite{resnet}, and DenseNet-121 and DenseNet-161 \cite{densenet}) considered in each case. For each dataset-backbone combination, we applied the Bayesian hyperparameter tuning regime of \cite{willard2024looks} for 72 GPU hours and used the best model found in terms of validation accuracy after projection as a reference ProtoPNet. For a full description of how these ProtoPNets were trained, see Appendix \ref{supp:training_ref_protopnets}.

We first evaluate the ability of \ourmethod{} to quickly produce a set of strong ProtoPNets. Across the eighteen settings described above, we measure the actual runtime required to produce a \ourmethod{} given a trained reference ProtoPNet. Appendix \ref{supp:hardware_details} describes the hardware used in this experiment. As shown in Figure \ref{fig:rset-comp-time}, we find that \textbf{a \ourmethod{} can be computed in less than 20 minutes across all eighteen settings considered}. This represents a negligible time cost compared to the cost of training one ProtoPNet.

Given that a \ourmethod{} can be produced in minutes, we next validate that \ourmethod{} quickly produces models that are accurate and meet user constraints. In each of the following sections, we start with a well trained ProtoPNet and iteratively remove up to $100$ random prototypes. If we find that no protoype can be removed from the model while remaining in the Rashomon set, we stop this procedure early.

We consider four baselines in the following experiments: naive prototype removal, where all last-layer coefficients for each removed prototype are simply set to $0$; naive prototype removal with retraining, where a similar removal procedure is applied and the last-layer of the ProtoPNet is retrained with an $\ell_1$ penalty on removed prototype weights; hard removal, where we strictly remove target prototypes and reoptimize all other last layer weights; and ProtoPDebug \cite{bontempelli2023concept}. Note that we only evaluate ProtoPDebug at 0, 25, 75, and 100 removals due to its long runtime.

\begin{figure*}[t!]
    \centering
    \includegraphics[width=1.0\linewidth]{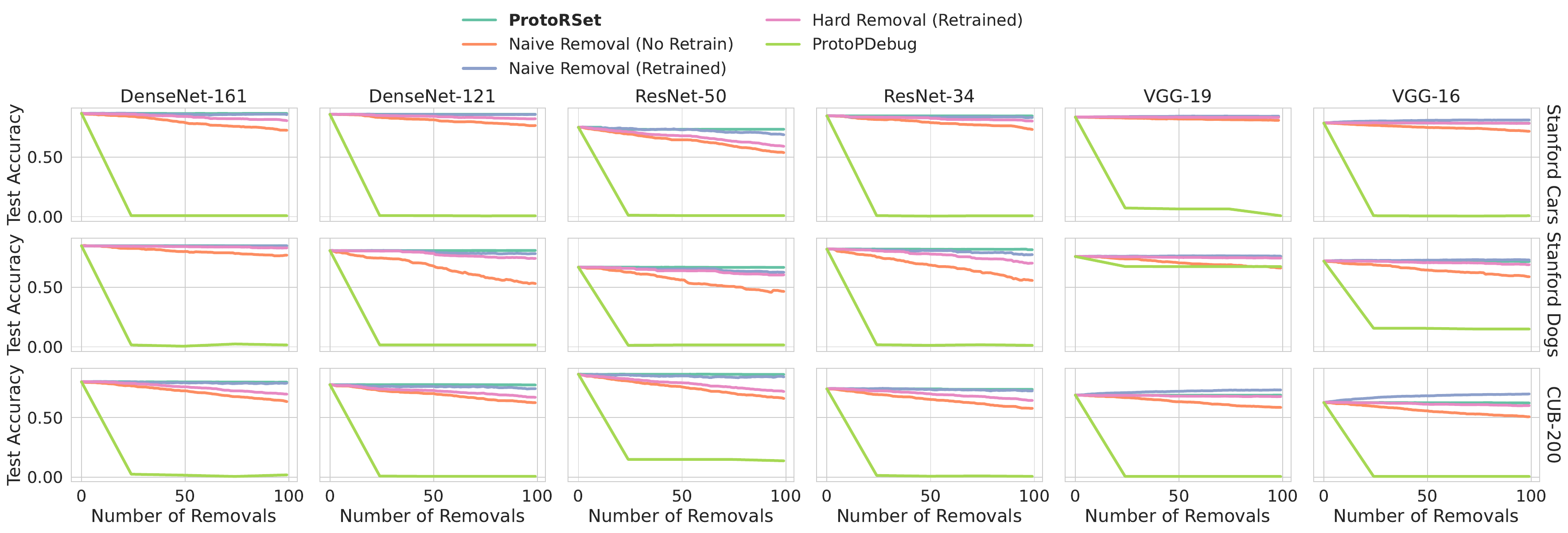}
    \caption{Change in test accuracy as random prototypes are removed. In all cases, we see that removing prototypes using ProtoRSet maintains or slightly improves the accuracy of the original model. Only naive removal with retraining maintains comparable accuracy.}
    \label{fig:accuracy_fig}
\end{figure*}

\textbf{\ourmethod{} Produces Accurate Models.} Figure \ref{fig:accuracy_fig} presents the test accuracy of each model produced in this experiment as a function of the number of prototypes removed. We find that, across all six backbones and all three datasets, \textbf{\ourmethod{} maintains test accuracy as constraints are added.} In contrast, every method except hard removal shows decreased accuracy as more random prototypes are removed.  

\begin{figure*}[t]
    \centering
    \includegraphics[width=1.0\linewidth]{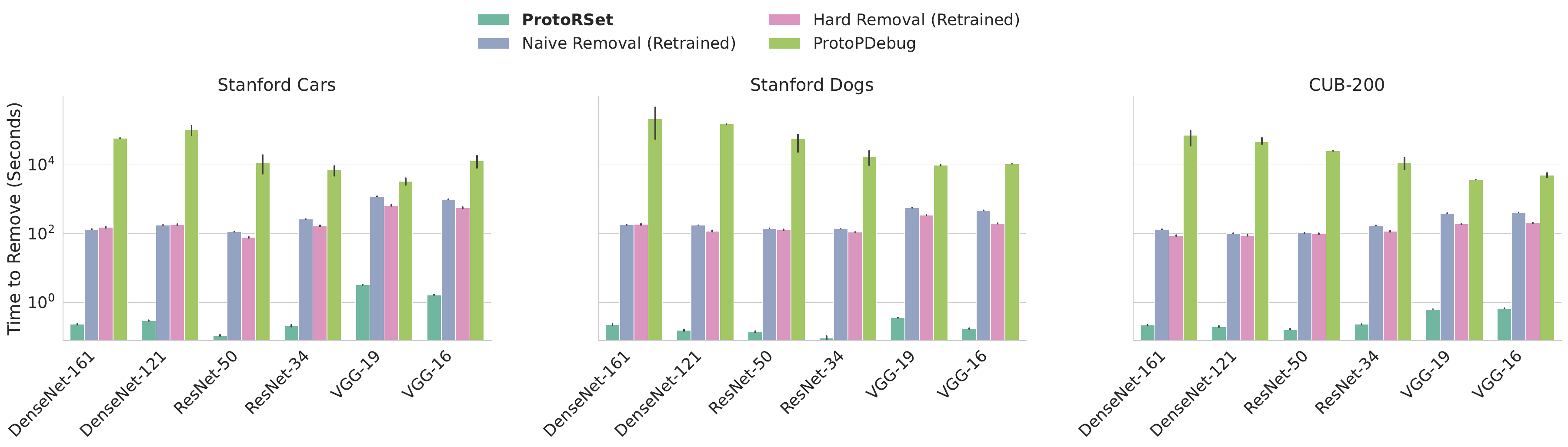}
    \caption{Time in seconds required to remove a single prototype, averaged over 100 iterations of removal. In all cases, ProtoRSet removes prototypes almost instantly. In contrast, removing a prototype then retraining the last layer can take orders of magnitude longer. We exclude naive removal without retraining because it is simply updating a value in an array, and as such is nearly instantaneous.}
    \label{fig:timing_fig}
\end{figure*}

\textbf{\ourmethod{} is Fast.} Figure \ref{fig:timing_fig} presents the time required to remove a prototype using \ourmethod{} versus each baseline except naive removal without retraining. We observe that, across all backbones and datasets, \ourmethod{} removes prototypes orders of magnitude faster than each baseline method. In fact, \ourmethod{} never requires more than a few seconds to produce a model satisfying new user constraints, making prototype removal a viable component of a real-time user interface. Naive removal without retraining tends to be faster, but at the cost of substantial decreases in accuracy.

\begin{figure*}[t]
    \centering
    \includegraphics[width=1.0\linewidth]{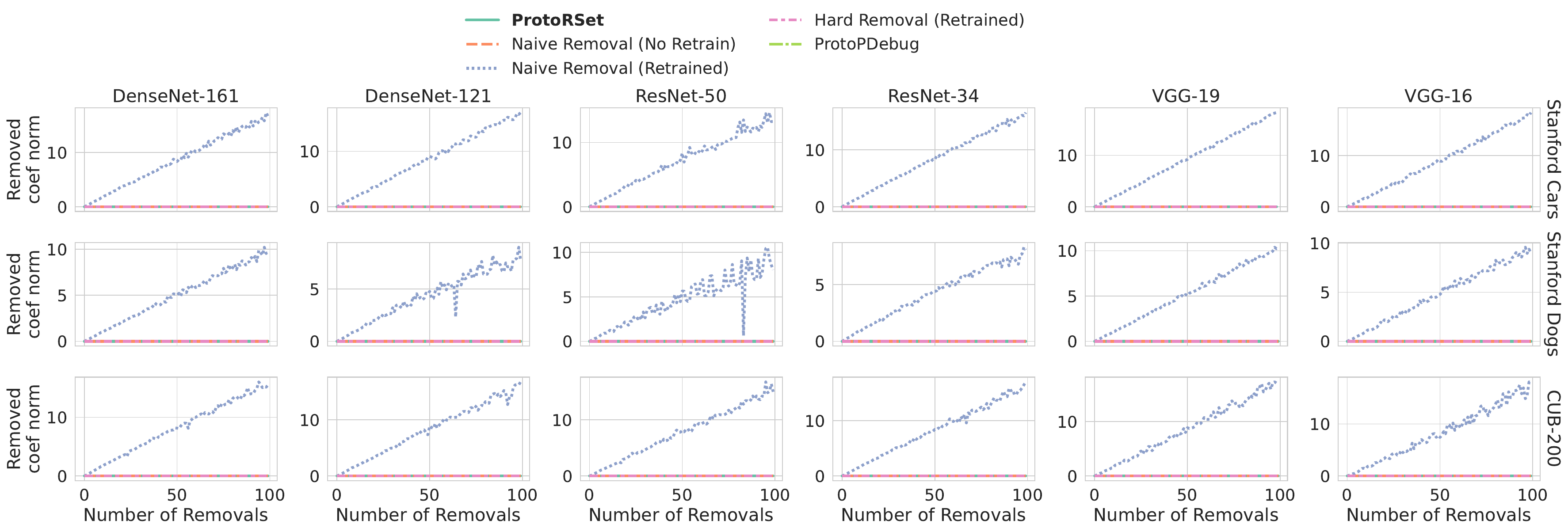}
    \caption{The $\ell_1$ norm of all coefficients associated with a ``removed'' prototype. Here, a larger $\ell_1$ norm indicates that the model is not meeting user constraints, since ``removed'' prototypes are still receiving a large weight in the model. Note that ProtoRSet and naive removal without retraining consistently produce models with an $\ell_1$ norm of approximately $0$, resulting in overlapping lines. In contrast, naive prototype removal with retraining does not guarantee that prototypes remain removed, resulting in $\ell_1$ norms that increase with the number of prototypes we attempt to remove.}
    \label{fig:removal_success}
\end{figure*}

\textbf{\ourmethod{} Guarantees Constraints are Met.} Figure \ref{fig:removal_success} presents the $\ell_1$ norm of all coefficients corresponding to removed prototypes. As shown in the figure, \textbf{\ourmethod{} guarantees that constraints imposed by the user are met}. On the other hand, naive removal of prototypes with retraining does not guarantee that the given constraints are met, with the ``removed'' prototypes continuing to play a role in the models this method produces. This is because retraining applies these constraints using a loss term, meaning it is not guaranteed that they are met.

\subsection{User Study: Removing Synthetic Confounders Quickly}
\begin{figure}
    \centering
    \includegraphics[width=1.0\linewidth]{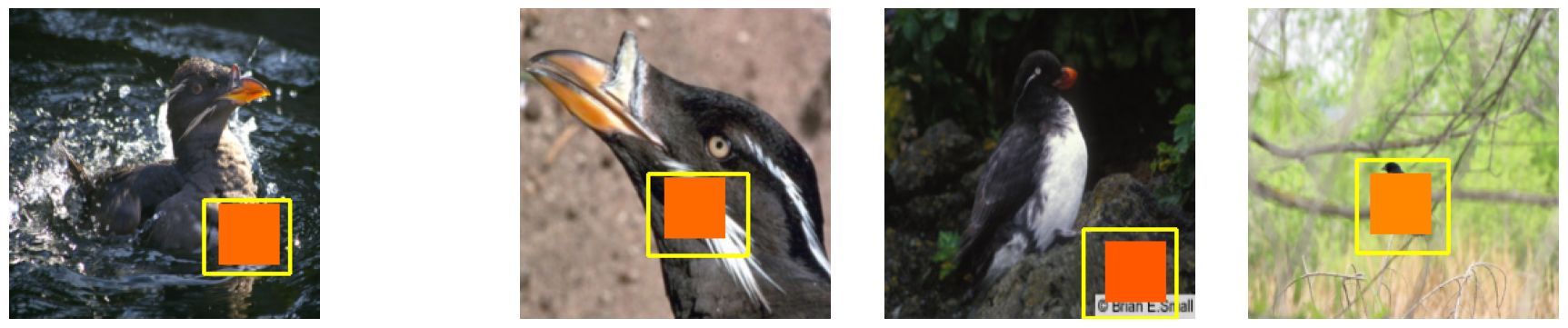}
    \caption{The four most activated images for a confounded prototype from the user study. This prototype has learned to focus on the synthetically added orange patch.}
    \label{fig:confounded_example}
\end{figure}

We demonstrate that laypeople can use \ourmethod{} to quickly remove confounding from a ProtoPNet. We added synthetic confounding to the training split of the CUB200 dataset and trained a ProtoPNet model using this corrupted data. In Figure \ref{fig:confounded_example}, we show prototypes that depend on the confounding orange squares added to the Rhinoceros Auklet class. With \ourmethod{}, users were able to identify and remove prototypes that depend on these synthetic confounders, producing a corrected model with only 2.1 minutes of computation on average.
Compared to the previous state of the art, \textbf{\ourmethod{} produced models with similar accuracy in roughly one fiftieth of the time}.

To create a confounded model, we modified the CUB200 training dataset such that a model might classify images using the ``wrong'' features. The CUB200 dataset contains 200 classes, each a different bird species. For the first 100 classes of the training set, we added a colored square to each image where each different color corresponded to a different bird class. The last 100 classes were untouched. We then trained a ProtoPNet with a VGG-19 backbone on this corrupted dataset. After training, we visually confirmed that this model had learned to reason using confounded prototypes, as shown in Figure \ref{fig:confounded_example}. For more detail on this procedure, see Appendix \ref{supp:confounding_details}.

Having trained the confounded model, we recruited 31 participants from the crowd-sourcing platform Prolific to remove the confounded prototypes from the model. Users were instructed to remove each prototype that focused entirely on one of the added color patches as quickly as possible using a simple user interface that allowed them to view and remove prototypes via \ourmethod{}, shown in Appendix Figure \ref{fig:ui_screenshot}. 

We compare to three baseline methods in this setting: ProtoPDebug, and naive prototype removal with and without retraining. We ran each method (\ourmethod{} and the three baselines) over the set of prototypes that each user removed. We measured the overall time spent on computation (i.e., not including time spent by the user looking at images). For each baseline method, we collect all prototypes removed by each user and remove them in a single pass. 
Note that this is a generous assumption for competitors -- whereas the runtime for \ourmethod{} includes the time for many refinement calls each time the user updates the model, we only measure the time for a single refinement call for ProtoPDebug and naive retraining. 
Additionally, we computed the change in validation accuracy after removing each user's specified prototypes using each method. For ProtoPDebug, we measured the time to remove a set of prototypes as the time taken for a training run to reach its maximum validation accuracy.

Table \ref{tab:ui_results} presents the results of this analysis. Using this interface, we found that users identified and removed an average of 80.8 confounded prototypes and produced a corrected model in an average of 33.6 minutes. Using real user feedback, we find that \textbf{\ourmethod{} meets user constraints in roughly one fiftieth the time taken by ProtoPDebug}. Moreover, \ourmethod{} preserves accuracy more reliably than ProtoPDebug, since ProtoPDebug sometimes produces models with substantially lower accuracy. In Appendix \ref{supp:debug_fails_feedback}, we show that, unlike ProtoPDebug, \ourmethod{} guarantees that user feedback is met. 

\begin{table}[]
    \centering
    \begin{tabular}{c||c|c|c}
        Method & \# Removed & Time (Mins)& $\Delta$ Acc \%\\
        \hline
        Debug & $80.8 \pm 44.6$ & $93.7 \pm 52.4$ & $+0.8 \pm 10.4$ \\
        Naive& $80.8 \pm 44.6$ & $0.0 \pm 0.0$ & $-6.4 \pm 2.9$\\
        Naive(T)& $80.8 \pm 44.6$ & $8.4 \pm 0.2$ & $-0.6 \pm 0.6$\\
        \textbf{Ours} & $80.8 \pm 44.6$ & $2.1 \pm 1.3$ & $-0.5 \pm 0.7$ 

    \end{tabular}
    \caption{User study results. Given the prototypes each user requests to remove, we measure mean and standard devation of the time required to remove each prototype and the change in validation accuracy from the initial ProtoPNet to the resulting model. ``Debug'' is ProtoPDebug, ``Naive (T)'' is the naive removal baseline with retraining, ``Naive'' is naive removal without retraining, and ``Ours'' is \ourmethod.}
    \label{tab:ui_results}
\end{table}

\subsection{Case Study: Refining a Skin Cancer Classification Model}
\label{subsec:skin_cancer}
\begin{figure}
    \centering
    \includegraphics[width=0.95\linewidth]{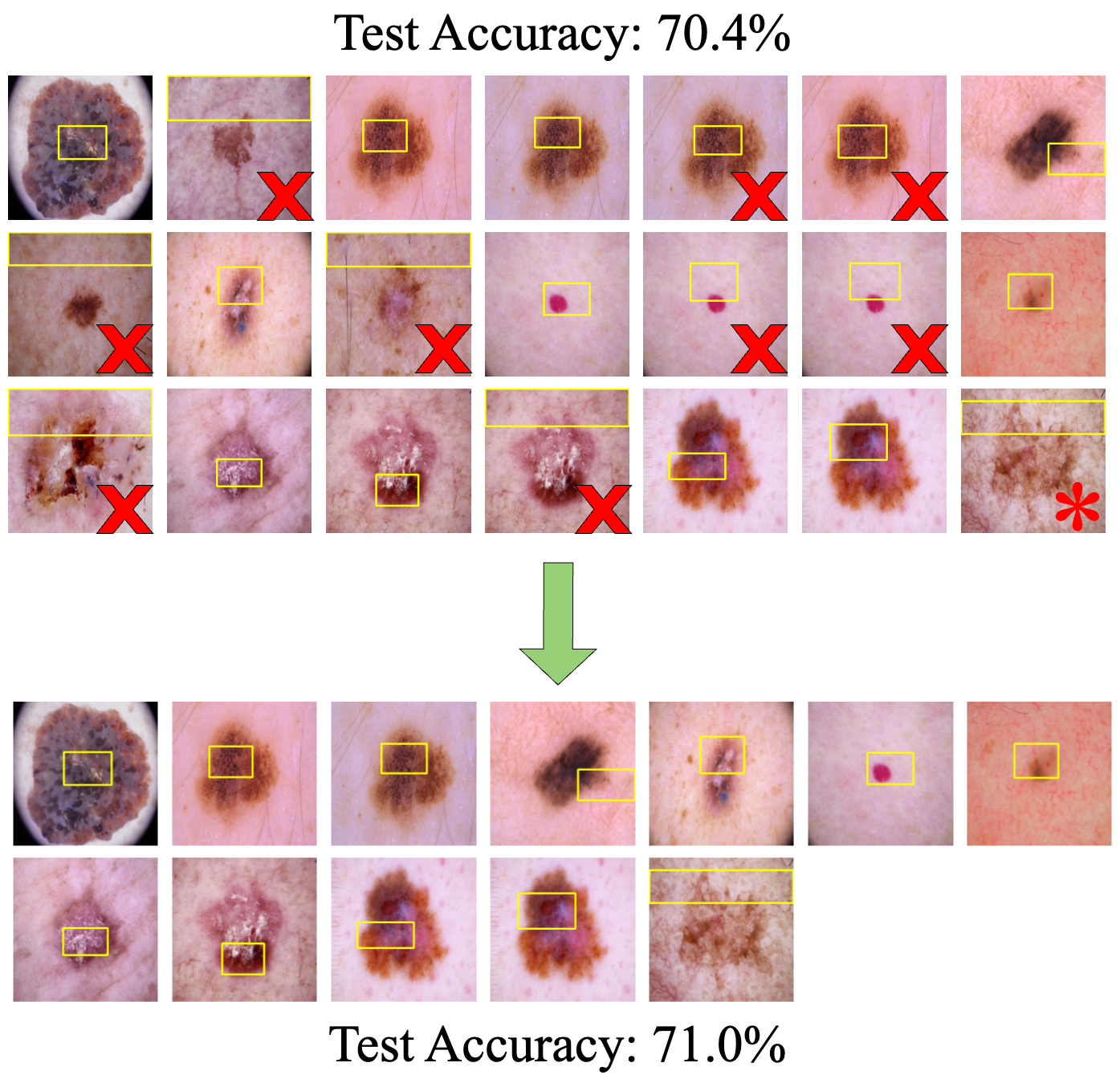}
    \caption{All prototypes from a ProtoPNet trained for skin lesion classification before (top) and after (bottom) refinement using \ourmethod. Removed prototypes are marked with a red ``X''. We attempted to remove the prototype marked with a red ``*,'' but \ourmethod{} correctly identified that this prototype could not be removed without a substantial loss in accuracy.}
    \label{fig:ham-case-study}
\end{figure}

When ProtoPNets are applied to medical tasks, they frequently encounter issues of prototype duplication and activation on medically irrelevant areas of the image \cite{barnett2021case}. We provide an example of \ourmethod{} in a realistic, high-stakes medical setting: skin cancer classification using the HAM10000 dataset \cite{tschandl2018ham10000}. By refining the resulting model using \ourmethod, \textbf{we reduced the total number of prototypes used by the model from 21 to 12 while increasing test accuracy from 70.4\% to 71.0\%} as shown in Figure \ref{fig:ham-case-study}.

HAM10000 consists of 11,720 skin lesion images, each labeled as one of seven lesion categories that include benign and malignant classifications. 
We trained a reference ProtoPNet on HAM10000 with a ResNet-34 backbone using the Bayesian hyperparameter tuning from \cite{willard2024looks} for 48 GPU hours, and selected the best model based on validation accuracy. We used \ourmethod{} to refine this model.

Out of 21 unique prototypes, we identified 10 prototypes that either did not focus on the lesion, or duplicated the reasoning of another prototype. We removed 9 of these prototypes. When attempting to remove the final seemingly confounded prototype in Figure \ref{fig:ham-case-study}, our \ourmethod{} reported that it was not possible to remove this prototype while remaining in the Rashomon set. We evaluated the accuracy of this claim by manually removing this prototype from the reference ProtoPNet, finding the \ourmethod{} with respect to the modified model, and repeating the removals specified above. We found that \textbf{ignoring \ourmethod's{} warning decreased test accuracy from 70.4\% to 57.9\% in the resulting model}.  This highlights an advantage of \ourmethod: when a user tries to impose a constraint that is strongly contradicted by the data, \ourmethod{} can directly identify this and prevent unexpected losses in model performance. \ourmethod{} produced a sparser, more accurate model even under non-expert refinement.



\section{Conclusion}
We introduced \ourmethod, a framework that computes a useful sampling from the Rashomon set of ProtoPNets. We showed that, across multiple datasets and backbone architectures, \ourmethod{} consistently produces models that meet all user constraints and maintain accuracy in a matter of seconds. Through a user study, we showed that this enables users to rapidly refine models, and that the accuracy of models from \ourmethod{} holds even under feedback from real users. Finally, we demonstrated the real-world utility of \ourmethod{} through a case study on skin lesion classification, where we demonstrated both that \ourmethod{} can improve accuracy given reasonable user feedback and that \ourmethod{} identifies unreasonable changes that \textit{cannot} be made while maintaining accuracy.

It is worth noting that, although we focused on simple ProtopNets leveraging cosine similarity in this work, \ourmethod{} is immediately compatible with many extensions of the original ProtoPNet \cite{chen2019looks}. Any method that modifies the backbone, the prototype layer, or the loss terms used by ProtoPNet is immediately applicable to \ourmethod{} (e.g., \cite{donnelly2022deformable, wang2021interpretable, yang2024fpn,barnett2021case, wei2024mprotonet, wang2023learning, nauta2023pip, ma2024looks}), as long as the final prediction head is a softmax over a linear layer of prototype similarities. \ourmethod{} is also applicable to any future extensions of ProtoPNet that address standing concerns around the interpretability of ProtoPNets \cite{huang2023evaluation, hoffmann2021looks}.

In this work, we set out to solve the deeply challenging problem of interacting with the Rashomon set of ProtoPNets. As such, it is natural that our solution comes with several limitations. First, \ourmethod{} is not the entire Rashomon set of ProtoPNets; this means that prior work studying the Rashomon set and its applications may not be immediately applicable to \ourmethod. Additionally, while \ourmethod{} is able to easily require prototypes be used with at least a given weight, doing so makes future operations with \ourmethod{} difficult because the resulting object is no longer an ellipsoid. 

Nonetheless, \ourmethod{} unlocks a new degree of usability for ProtoPNets. Never before has it been possible to debug and refine a ProtoPNet in real time, with guaranteed performance; thanks to \ourmethod, it is now.

\section{Acknowledgements}
We acknowledge funding from the National Science Foundation under 
grants HRD-2222336 and OIA-2218063, and the Department of Energy under grant DE-SC0023194. Additionally, this material is based upon work supported by the National Science
Foundation Graduate Research Fellowship under Grant No. DGE 2139754.
{
    \small
    \bibliographystyle{ieeenat_fullname}
    \bibliography{main}

\begin{thebibliography}{56}
\providecommand{\natexlab}[1]{#1}
\providecommand{\url}[1]{\texttt{#1}}
\expandafter\ifx\csname urlstyle\endcsname\relax
  \providecommand{\doi}[1]{doi: #1}\else
  \providecommand{\doi}{doi: \begingroup \urlstyle{rm}\Url}\fi

\bibitem[Barnett et~al.(2021)Barnett, Schwartz, Tao, Chen, Ren, Lo, and Rudin]{barnett2021case}
Alina~Jade Barnett, Fides~Regina Schwartz, Chaofan Tao, Chaofan Chen, Yinhao Ren, Joseph~Y Lo, and Cynthia Rudin.
\newblock {A Case-Based Interpretable Deep Learning Model for Classification of Mass Lesions in Digital Mammography}.
\newblock \emph{Nature Machine Intelligence}, 3\penalty0 (12):\penalty0 1061--1070, 2021.

\bibitem[Barnett et~al.(2024)Barnett, Guo, Jing, Ge, Kaplan, Kong, Karakis, Herlopian, Jayagopal, Taraschenko, et~al.]{barnett2024improving}
Alina~Jade Barnett, Zhicheng Guo, Jin Jing, Wendong Ge, Peter~W Kaplan, Wan~Yee Kong, Ioannis Karakis, Aline Herlopian, Lakshman~Arcot Jayagopal, Olga Taraschenko, et~al.
\newblock {Improving Clinician Performance in Classifying EEG Patterns on the Ictal--Interictal Injury Continuum Using Interpretable Machine Learning}.
\newblock \emph{NEJM AI}, 1\penalty0 (6):\penalty0 AIoa2300331, 2024.

\bibitem[B{\"o}hle et~al.(2022)B{\"o}hle, Fritz, and Schiele]{bohle2022b}
Moritz B{\"o}hle, Mario Fritz, and Bernt Schiele.
\newblock {B-Cos Networks: Alignment Is All We Need for Interpretability}.
\newblock In \emph{Proceedings of the IEEE/CVF Conference on Computer Vision and Pattern Recognition}, pages 10329--10338, 2022.

\bibitem[Boner et~al.(2024)Boner, Chen, Semenova, Parr, and Rudin]{boner2024using}
Zachery Boner, Harry Chen, Lesia Semenova, Ronald Parr, and Cynthia Rudin.
\newblock {Using Noise to Infer Aspects of Simplicity Without Learning}.
\newblock In \emph{The Thirty-eighth Annual Conference on Neural Information Processing Systems}, 2024.

\bibitem[Bontempelli et~al.(2023)Bontempelli, Teso, Tentori, Giunchiglia, Passerini, et~al.]{bontempelli2023concept}
Andrea Bontempelli, Stefano Teso, Katya Tentori, Fausto Giunchiglia, Andrea Passerini, et~al.
\newblock {Concept-Level Debugging of Part-Prototype Networks}.
\newblock In \emph{Proceedings of the The Eleventh International Conference on Learning Representations (ICLR 23)}. ICLR 2023, 2023.

\bibitem[Breiman(2001)]{breiman2001statistical}
Leo Breiman.
\newblock {Statistical Modeling: The Two Cultures (With Comments and a Rejoinder by the Author)}.
\newblock \emph{Statistical science}, 16\penalty0 (3):\penalty0 199--231, 2001.

\bibitem[Chen et~al.(2019)Chen, Li, Tao, Barnett, Rudin, and Su]{chen2019looks}
Chaofan Chen, Oscar Li, Daniel Tao, Alina Barnett, Cynthia Rudin, and Jonathan~K Su.
\newblock {This Looks Like That: Deep Learning for Interpretable Image Recognition}.
\newblock \emph{Advances in Neural Information Processing Systems}, 32, 2019.

\bibitem[Choukali et~al.(2024)Choukali, Amirani, Valizadeh, Abbasi, and Komeili]{choukali2024pseudo}
Mohammad~Amin Choukali, Mehdi~Chehel Amirani, Morteza Valizadeh, Ata Abbasi, and Majid Komeili.
\newblock {Pseudo-Class Part Prototype Networks for Interpretable Breast Cancer Classification}.
\newblock \emph{Scientific Reports}, 14\penalty0 (1):\penalty0 10341, 2024.

\bibitem[Dong and Rudin(2020)]{dong2019variable}
Jiayun Dong and Cynthia Rudin.
\newblock {Exploring the Cloud of Variable Importance for the Set of All Good Models}.
\newblock \emph{Nature Machine Intelligence}, 2\penalty0 (12):\penalty0 810--824, 2020.

\bibitem[Donnelly et~al.(2022)Donnelly, Barnett, and Chen]{donnelly2022deformable}
Jon Donnelly, Alina~Jade Barnett, and Chaofan Chen.
\newblock {Deformable Protopnet: An Interpretable Image Classifier Using Deformable Prototypes}.
\newblock In \emph{Proceedings of the IEEE/CVF Conference on Computer Vision and Pattern Recognition}, pages 10265--10275, 2022.

\bibitem[Donnelly et~al.(2023)Donnelly, Katta, Rudin, and Browne]{donnelly2023rashomon}
Jon Donnelly, Srikar Katta, Cynthia Rudin, and Edward Browne.
\newblock {The {R}ashomon importance distribution: Getting rid of unstable, single model-based variable importance}.
\newblock \emph{Advances in Neural Information Processing Systems}, 36:\penalty0 6267--6279, 2023.

\bibitem[Doshi-Velez and Kim(2017)]{doshi2017towards}
Finale Doshi-Velez and Been Kim.
\newblock {Towards a Rigorous Science of Interpretable Machine Learning}.
\newblock \emph{arXiv preprint arXiv:1702.08608}, 2017.

\bibitem[Fisher et~al.(2019)Fisher, Rudin, and Dominici]{fisher2019all}
Aaron Fisher, Cynthia Rudin, and Francesca Dominici.
\newblock {All Models Are Wrong, but Many Are Useful: Learning a Variable's Importance by Studying an Entire Class of Prediction Models Simultaneously}.
\newblock \emph{Journal of Machine Learning Research}, 20\penalty0 (177):\penalty0 1--81, 2019.

\bibitem[Food and Administration(2021)]{us2021artificial}
US Food and Drug Administration.
\newblock {Machine Learning (AI/ML)-Based Software as a Medical Device (SaMD) Action Plan}.
\newblock \emph{US Food \& Drug Administration: Silver Spring, MD, USA}, 2021.

\bibitem[for Official Publications of~the European~Communities(2021)]{com2021laying}
Office for Official Publications of~the European~Communities.
\newblock {Laying Down Harmonised Rules on Artificial Intelligence (Artificial Intelligence Act) and Amending Certain Union Legislative Acts}.
\newblock \emph{Proposal for a regulation of the European parliament and of the council}, 2021.

\bibitem[Geis et~al.(2019)Geis, Brady, Wu, Spencer, Ranschaert, Jaremko, Langer, Borondy~Kitts, Birch, Shields, et~al.]{geis2019ethics}
J~Raymond Geis, Adrian~P Brady, Carol~C Wu, Jack Spencer, Erik Ranschaert, Jacob~L Jaremko, Steve~G Langer, Andrea Borondy~Kitts, Judy Birch, William~F Shields, et~al.
\newblock {Ethics of Artificial Intelligence in Radiology: Summary of the Joint European and North American Multisociety Statement}.
\newblock \emph{Radiology}, 293\penalty0 (2):\penalty0 436--440, 2019.

\bibitem[He et~al.(2016)He, Zhang, Ren, and Sun]{resnet}
Kaiming He, Xiangyu Zhang, Shaoqing Ren, and Jian Sun.
\newblock Deep {Residual} {Learning} for {Image} {Recognition}.
\newblock In \emph{2016 {IEEE} {Conference} on {Computer} {Vision} and {Pattern} {Recognition} ({CVPR})}, pages 770--778, Las Vegas, NV, USA, 2016. IEEE.

\bibitem[Hoffmann et~al.(2021)Hoffmann, Fanconi, Rade, and Kohler]{hoffmann2021looks}
Adrian Hoffmann, Claudio Fanconi, Rahul Rade, and Jonas Kohler.
\newblock {This Looks Like That... Does it? Shortcomings of Latent Space Prototype Interpretability in Deep Networks}.
\newblock \emph{arXiv preprint arXiv:2105.02968}, 2021.

\bibitem[Hsu and Calmon(2022)]{hsu2022rashomon}
Hsiang Hsu and Flavio Calmon.
\newblock {Rashomon Capacity: A Metric for Predictive Multiplicity in Classification}.
\newblock \emph{Advances in Neural Information Processing Systems}, 35:\penalty0 28988--29000, 2022.

\bibitem[Huang et~al.(2017)Huang, Liu, Van Der~Maaten, and Weinberger]{densenet}
Gao Huang, Zhuang Liu, Laurens Van Der~Maaten, and Kilian~Q. Weinberger.
\newblock Densely {Connected} {Convolutional} {Networks}.
\newblock In \emph{2017 {IEEE} {Conference} on {Computer} {Vision} and {Pattern} {Recognition} ({CVPR})}, pages 2261--2269. IEEE, 2017.

\bibitem[Huang et~al.(2023)Huang, Xue, Huang, Zhang, Song, Jing, and Song]{huang2023evaluation}
Qihan Huang, Mengqi Xue, Wenqi Huang, Haofei Zhang, Jie Song, Yongcheng Jing, and Mingli Song.
\newblock {Evaluation and Improvement of Interpretability for Self-Explainable Part-Prototype Networks}.
\newblock In \emph{{Proceedings of the IEEE/CVF International Conference on Computer Vision}}, pages 2011--2020, 2023.

\bibitem[Khosla et~al.(2011)Khosla, Jayadevaprakash, Yao, and Li]{khosla2011novel}
Aditya Khosla, Nityananda Jayadevaprakash, Bangpeng Yao, and Fei-Fei Li.
\newblock {Novel Dataset for Fine-Grained Image Categorization: Stanford Dogs}.
\newblock In \emph{Proc. CVPR Workshop on Fine-Grained Visual Categorization (FGVC)}, 2011.

\bibitem[Koh et~al.(2020)Koh, Nguyen, Tang, Mussmann, Pierson, Kim, and Liang]{koh2020concept}
Pang~Wei Koh, Thao Nguyen, Yew~Siang Tang, Stephen Mussmann, Emma Pierson, Been Kim, and Percy Liang.
\newblock {Concept Bottleneck Models}.
\newblock In \emph{International Conference on Machine Learning}, pages 5338--5348. PMLR, 2020.

\bibitem[Krause et~al.(2013)Krause, Stark, Deng, and Fei-Fei]{KrauseStarkDengFei-Fei_3DRR2013}
Jonathan Krause, Michael Stark, Jia Deng, and Li Fei-Fei.
\newblock {3D Object Representations for Fine-Grained Categorization}.
\newblock In \emph{4th International IEEE Workshop on 3D Representation and Recognition (3dRR-13)}, Sydney, Australia, 2013.

\bibitem[Kulynych et~al.(2023)Kulynych, Hsu, Troncoso, and Calmon]{kulynych2023arbitrary}
Bogdan Kulynych, Hsiang Hsu, Carmela Troncoso, and Flavio~P Calmon.
\newblock {Arbitrary Decisions Are a Hidden Cost of Differentially Private Training}.
\newblock In \emph{Proceedings of the 2023 ACM Conference on Fairness, Accountability, and Transparency}, pages 1609--1623, 2023.

\bibitem[Kurzhanskiy and Varaiya(2006)]{kurzhanskiy2006ellipsoidal}
Alex~A Kurzhanskiy and Pravin Varaiya.
\newblock {Ellipsoidal Toolbox (ET)}.
\newblock In \emph{Proceedings of the 45th IEEE Conference on Decision and Control}, pages 1498--1503. IEEE, 2006.

\bibitem[Li et~al.()Li, Netzorg, Cheng, Zhang, and Yu]{liimproving}
Aaron~Jiaxun Li, Robin Netzorg, Zhihan Cheng, Zhuoqin Zhang, and Bin Yu.
\newblock {Improving Prototypical Visual Explanations With Reward Reweighing, Reselection, and Retraining}.
\newblock In \emph{Forty-first International Conference on Machine Learning}.

\bibitem[Li et~al.(2018)Li, Liu, Chen, and Rudin]{li2018deep}
Oscar Li, Hao Liu, Chaofan Chen, and Cynthia Rudin.
\newblock {Deep Learning for Case-Based Reasoning Through Prototypes: A Neural Network That Explains Its Predictions}.
\newblock In \emph{Proceedings of the {AAAI} Conference on Artificial Intelligence}, 2018.

\bibitem[Liu et~al.(2022)Liu, Zhong, Li, Seltzer, and Rudin]{liu2022fasterrisk}
Jiachang Liu, Chudi Zhong, Boxuan Li, Margo Seltzer, and Cynthia Rudin.
\newblock {FasterRisk: Fast and Accurate Interpretable Risk Scores}.
\newblock \emph{Advances in Neural Information Processing Systems}, 35:\penalty0 17760--17773, 2022.

\bibitem[Ma et~al.(2024)Ma, Zhao, Chen, and Rudin]{ma2024looks}
Chiyu Ma, Brandon Zhao, Chaofan Chen, and Cynthia Rudin.
\newblock {This Looks Like Those: Illuminating Prototypical Concepts Using Multiple Visualizations}.
\newblock \emph{Advances in Neural Information Processing Systems}, 36, 2024.

\bibitem[Marx et~al.(2020)Marx, Calmon, and Ustun]{marx2020predictive}
Charles Marx, Flavio Calmon, and Berk Ustun.
\newblock {Predictive Multiplicity in Classification}.
\newblock In \emph{International Conference on Machine Learning}, pages 6765--6774. PMLR, 2020.

\bibitem[Nauta et~al.(2021{\natexlab{a}})Nauta, Jutte, Provoost, and Seifert]{nauta2021looks}
Meike Nauta, Annemarie Jutte, Jesper Provoost, and Christin Seifert.
\newblock {This Looks Like That, Because... Explaining Prototypes for Interpretable Image Recognition}.
\newblock In \emph{Joint European Conference on Machine Learning and Knowledge Discovery in Databases}, pages 441--456. Springer, 2021{\natexlab{a}}.

\bibitem[Nauta et~al.(2021{\natexlab{b}})Nauta, Van~Bree, and Seifert]{nauta2021neural}
Meike Nauta, Ron Van~Bree, and Christin Seifert.
\newblock {Neural Prototype Trees for Interpretable Fine-Grained Image Recognition}.
\newblock In \emph{Proceedings of the IEEE/CVF conference on Computer Vision and Pattern Recognition}, pages 14933--14943, 2021{\natexlab{b}}.

\bibitem[Nauta et~al.(2023)Nauta, Schl{\"o}tterer, Van~Keulen, and Seifert]{nauta2023pip}
Meike Nauta, J{\"o}rg Schl{\"o}tterer, Maurice Van~Keulen, and Christin Seifert.
\newblock {Pip-Net: Patch-Based Intuitive Prototypes for Interpretable Image Classification}.
\newblock In \emph{Proceedings of the IEEE/CVF Conference on Computer Vision and Pattern Recognition}, pages 2744--2753, 2023.

\bibitem[Rudin et~al.(2022)Rudin, Chen, Chen, Huang, Semenova, and Zhong]{rudin2022interpretable}
Cynthia Rudin, Chaofan Chen, Zhi Chen, Haiyang Huang, Lesia Semenova, and Chudi Zhong.
\newblock {Interpretable Machine Learning: Fundamental Principles and 10 Grand Challenges}.
\newblock \emph{Statistic Surveys}, 16:\penalty0 1--85, 2022.

\bibitem[Rudin et~al.(2024)Rudin, Zhong, Semenova, Seltzer, Parr, Liu, Katta, Donnelly, Chen, and Boner]{rudin2024amazing}
Cynthia Rudin, Chudi Zhong, Lesia Semenova, Margo Seltzer, Ronald Parr, Jiachang Liu, Srikar Katta, Jon Donnelly, Harry Chen, and Zachery Boner.
\newblock {Amazing Things Come From Having Many Good Models}.
\newblock In \emph{Proceedings of the International Conference on Machine Learning}, 2024.

\bibitem[Rymarczyk et~al.(2021)Rymarczyk, Struski, Tabor, and Zieli{\'n}ski]{rymarczyk2021protopshare}
Dawid Rymarczyk, {\L}ukasz Struski, Jacek Tabor, and Bartosz Zieli{\'n}ski.
\newblock {{ProtoPShare}: Prototypical Parts Sharing for Similarity Discovery in Interpretable Image Classification}.
\newblock In \emph{Proceedings of the 27th ACM SIGKDD Conference on Knowledge Discovery \& Data Mining}, pages 1420--1430, 2021.

\bibitem[Rymarczyk et~al.(2022)Rymarczyk, Struski, G{\'o}rszczak, Lewandowska, Tabor, and Zieli{\'n}ski]{rymarczyk2022interpretable}
Dawid Rymarczyk, {\L}ukasz Struski, Micha{\l} G{\'o}rszczak, Koryna Lewandowska, Jacek Tabor, and Bartosz Zieli{\'n}ski.
\newblock {Interpretable Image Classification With Differentiable Prototypes Assignment}.
\newblock In \emph{European Conference on Computer Vision}, pages 351--368. Springer, 2022.

\bibitem[Rymarczyk et~al.(2023)Rymarczyk, van~de Weijer, Zieli{\'n}ski, and Twardowski]{rymarczyk2023icicle}
Dawid Rymarczyk, Joost van~de Weijer, Bartosz Zieli{\'n}ski, and Bartlomiej Twardowski.
\newblock {{ICICLE}: Interpretable Class Incremental Continual Learning}.
\newblock In \emph{Proceedings of the IEEE/CVF International Conference on Computer Vision}, pages 1887--1898, 2023.

\bibitem[Semenova et~al.(2022)Semenova, Rudin, and Parr]{semenova2022existence}
Lesia Semenova, Cynthia Rudin, and Ronald Parr.
\newblock {On the Existence of Simpler Machine Learning Models}.
\newblock In \emph{Proceedings of the 2022 {ACM} Conference on Fairness, Accountability, and Transparency}, pages 1827--1858, 2022.

\bibitem[Semenova et~al.(2024)Semenova, Chen, Parr, and Rudin]{semenova2024path}
Lesia Semenova, Harry Chen, Ronald Parr, and Cynthia Rudin.
\newblock {A Path to Simpler Models Starts With Noise}.
\newblock \emph{Advances in Neural Information Processing Systems}, 36, 2024.

\bibitem[Simonyan and Zisserman(2015)]{vgg}
Karen Simonyan and Andrew Zisserman.
\newblock {Very Deep Convolutional Networks for Large-Scale Image Recognition}.
\newblock In \emph{Proceedings of the 3rd International Conference on Learning Representations (ICLR)}, 2015.

\bibitem[Smith et~al.(2020)Smith, Mansilla, and Goulding]{smith2020model}
Gavin Smith, Roberto Mansilla, and James Goulding.
\newblock {Model Class Reliance for Random Forests}.
\newblock \emph{Advances in Neural Information Processing Systems}, 33:\penalty0 22305--22315, 2020.

\bibitem[Taesiri et~al.(2022)Taesiri, Nguyen, and Nguyen]{taesiri2022visual}
Mohammad~Reza Taesiri, Giang Nguyen, and Anh Nguyen.
\newblock {Visual Correspondence-Based Explanations Improve AI Robustness and Human-Ai Team Accuracy}.
\newblock \emph{Advances in Neural Information Processing Systems}, 35:\penalty0 34287--34301, 2022.

\bibitem[Tschandl et~al.(2018)Tschandl, Rosendahl, and Kittler]{tschandl2018ham10000}
Philipp Tschandl, Cliff Rosendahl, and Harald Kittler.
\newblock {The HAM10000 Dataset, a Large Collection of Multi-Source Dermatoscopic Images of Common Pigmented Skin Lesions}.
\newblock \emph{Scientific data}, 5\penalty0 (1):\penalty0 1--9, 2018.

\bibitem[Wah et~al.(2011)Wah, Branson, Welinder, Perona, and Belongie]{WahCUB_200_2011}
C. Wah, S. Branson, P. Welinder, P. Perona, and S. Belongie.
\newblock The {C}altech-{UCSD} birds-200-2011 {D}ataset.
\newblock Technical Report CNS-TR-2011-001, California Institute of Technology, 2011.

\bibitem[Wang et~al.(2023)Wang, Liu, Chen, Liu, Tian, McCarthy, Frazer, and Carneiro]{wang2023learning}
Chong Wang, Yuyuan Liu, Yuanhong Chen, Fengbei Liu, Yu Tian, Davis McCarthy, Helen Frazer, and Gustavo Carneiro.
\newblock {Learning Support and Trivial Prototypes for Interpretable Image Classification}.
\newblock In \emph{Proceedings of the IEEE/CVF International Conference on Computer Vision}, pages 2062--2072, 2023.

\bibitem[Wang et~al.(2021)Wang, Liu, Wang, and Jing]{wang2021interpretable}
Jiaqi Wang, Huafeng Liu, Xinyue Wang, and Liping Jing.
\newblock {Interpretable Image Recognition by Constructing Transparent Embedding Space}.
\newblock In \emph{Proceedings of the IEEE/CVF International Conference on Computer Vision}, pages 895--904, 2021.

\bibitem[Watson-Daniels et~al.(2023{\natexlab{a}})Watson-Daniels, Barocas, Hofman, and Chouldechova]{watson2023multi}
Jamelle Watson-Daniels, Solon Barocas, Jake~M Hofman, and Alexandra Chouldechova.
\newblock {Multi-Target Multiplicity: Flexibility and Fairness in Target Specification Under Resource Constraints}.
\newblock In \emph{Proceedings of the 2023 ACM Conference on Fairness, Accountability, and Transparency}, pages 297--311, 2023{\natexlab{a}}.

\bibitem[Watson-Daniels et~al.(2023{\natexlab{b}})Watson-Daniels, Parkes, and Ustun]{watson2023predictive}
Jamelle Watson-Daniels, David~C Parkes, and Berk Ustun.
\newblock {Predictive Multiplicity in Probabilistic Classification}.
\newblock In \emph{Proceedings of the AAAI Conference on Artificial Intelligence}, pages 10306--10314, 2023{\natexlab{b}}.

\bibitem[Wei et~al.(2024)Wei, Tam, and Tang]{wei2024mprotonet}
Yuanyuan Wei, Roger Tam, and Xiaoying Tang.
\newblock {MProtoNet: A Case-Based Interpretable Model for Brain Tumor Classification With 3D Multi-Parametric Magnetic Resonance Imaging}.
\newblock In \emph{Medical Imaging with Deep Learning}, pages 1798--1812. PMLR, 2024.

\bibitem[Willard et~al.(2024)Willard, Moffett, Mokel, Donnelly, Guo, Yang, Kim, Barnett, and Rudin]{willard2024looks}
Frank Willard, Luke Moffett, Emmanuel Mokel, Jon Donnelly, Stark Guo, Julia Yang, Giyoung Kim, Alina~Jade Barnett, and Cynthia Rudin.
\newblock {This Looks Better Than That: Better Interpretable Models With {ProtoPNext}}.
\newblock \emph{arXiv preprint arXiv:2406.14675}, 2024.

\bibitem[Xin et~al.(2022)Xin, Zhong, Chen, Takagi, Seltzer, and Rudin]{xin2022exploring}
Rui Xin, Chudi Zhong, Zhi Chen, Takuya Takagi, Margo Seltzer, and Cynthia Rudin.
\newblock {Exploring the Whole Rashomon Set of Sparse Decision Trees}.
\newblock \emph{Advances in Neural Information Processing Systems}, 35:\penalty0 14071--14084, 2022.

\bibitem[Yang et~al.(2024)Yang, Barnett, Donnelly, Kishore, Fang, Schwartz, Chen, Lo, and Rudin]{yang2024fpn}
Julia Yang, Alina~Jade Barnett, Jon Donnelly, Satvik Kishore, Jerry Fang, Fides~Regina Schwartz, Chaofan Chen, Joseph~Y Lo, and Cynthia Rudin.
\newblock {{FPN-Iaia-Bl}: A Multi-Scale Interpretable Deep Learning Model for Classification of Mass Margins in Digital Mammography}.
\newblock In \emph{Proceedings of the IEEE/CVF Conference on Computer Vision and Pattern Recognition}, pages 5003--5009, 2024.

\bibitem[You et~al.(2023)You, Qu, Gatti, Jain, and Wong]{you2023sum}
Weiqiu You, Helen Qu, Marco Gatti, Bhuvnesh Jain, and Eric Wong.
\newblock {Sum-of-Parts Models: Faithful Attributions for Groups of Features}.
\newblock \emph{arXiv preprint arXiv:2310.16316}, 2023.

\bibitem[Zhong et~al.(2024)Zhong, Chen, Liu, Seltzer, and Rudin]{zhong2024exploring}
Chudi Zhong, Zhi Chen, Jiachang Liu, Margo Seltzer, and Cynthia Rudin.
\newblock {Exploring and Interacting With the Set of Good Sparse Generalized Additive Models}.
\newblock \emph{Advances in Neural Information Processing Systems}, 36, 2024.

\end{thebibliography}
}


\onecolumn
\maketitlesupplementary
\setcounter{page}{1}
\setcounter{section}{0}

\section{Sampling Unrestricted Last Layers from the Rashomon Set Without Explicitly Computing the Hessian}
\label{supp:sampling_no_hessian}

Let $\mathbf{z} \in \mathbb{R}^m$ denote a vector of prototype similarities, such that $h(\mathbf{z}) \in \mathbb{R}^t$ is the vector of predicted class probabilities for input $\mathbf{z}$. The Hessian matrix of $h$ with respect to $\mathbf{z}$ for a standard, multi-class logistic regression problem is given (written in terms of block matrices) as:
\begin{align*}
    \mathbf{H} &= - \begin{bmatrix}
        h_1(\mathbf{z}) (1 - h_1(\mathbf{z})) \mathbf{z} \mathbf{z}^T & 
        -h_1(\mathbf{z}) h_2(\mathbf{z}) \mathbf{z} \mathbf{z}^T & \hdots & -h_1(\mathbf{z}) h_t(\mathbf{z}) \mathbf{z} \mathbf{z}^T\\
        -h_2(\mathbf{z}) h_1(\mathbf{z}) \mathbf{z} \mathbf{z}^T &  h_2(\mathbf{z}) (1 - h_2(\mathbf{z})) \mathbf{z} \mathbf{z}^T &\hdots & -h_2(\mathbf{z}) h_t(\mathbf{z}) \mathbf{z} \mathbf{z}^T\\
        \vdots & \vdots & \ddots & \vdots\\
        -h_t(\mathbf{z}) h_1(\mathbf{z}) \mathbf{z} \mathbf{z}^T &  -h_t(\mathbf{z}) h_2(\mathbf{z}) \mathbf{z} \mathbf{z}^T &\hdots &  h_t(\mathbf{z}) (1 - h_t(\mathbf{z})) \mathbf{z} \mathbf{z}^T\\
    \end{bmatrix}\\
    &= \left(\mathbf{\Lambda} - h(\mathbf{z}) h(\mathbf{z})^T\right) \otimes \mathbf{z}\mathbf{z}^T,
\end{align*}
where
\begin{align*}
    \mathbf{\Lambda}_{ij} &:= \mathbf{1}[i=j]h_i(\mathbf{z}).
\end{align*}
In order to sample a member of a Rashomon set that falls along direction $\mathbf{d} \in \mathbb{R}^{mt},$ we need to compute a value $\tau$ such that
\begin{align*}
    \tau = \sqrt{\frac{2\kappa(\theta - \ell(\mathbf{w}^*_h))}{\mathbf{d}^T \mathbf{H} \mathbf{d}}},
\end{align*}
with each term defined as in Section \ref{subsec:interaction} of the main paper. The key computational constraint here is the operation $\mathbf{d}^T\mathbf{H}\mathbf{d},$ which involves the large Hessian matrix $\mathbf{H} \in \mathbb{R}^{mt \times mt}.$  However, the quantity $\mathbf{d}^T\mathbf{H}\mathbf{d}$ can be computed without explicitly storing $\mathbf{H}.$

Let $\text{mat}: \mathbb{R}^{mt} \to \mathbb{R}^{m \times t}$ denote an operation that reshapes a vector into a matrix, and let $\text{vec}: \mathbb{R}^{m \times t} \to \mathbb{R}^{mt}$ denote the inverse operaton, which reshapes a matrix into a vector such that $\text{vec}(\text{mat}(\mathbf{d})) = \mathbf{d}.$ We can then leverage the property of the Kronecker product that $(\mathbf{A} \otimes \mathbf{B}) \text{vec}(\mathbf{C}) = \text{vec}(\mathbf{B} \mathbf{C} \mathbf{A}^T)$ to compute:
\begin{align*}
    \mathbf{d}^T\mathbf{H}\mathbf{d} &= \mathbf{d}^T \left( \left(\mathbf{\Lambda} - h(\mathbf{z}) h(\mathbf{z})^T\right) \otimes \mathbf{z}\mathbf{z}^T \right) \mathbf{d}\\
    &= \mathbf{d}^T \left( \left(\mathbf{\Lambda} - h(\mathbf{z}) h(\mathbf{z})^T\right) \otimes \mathbf{z}\mathbf{z}^T \right) \text{vec}\left(\text{mat}\left(\mathbf{d}\right)\right)\\
    &= \mathbf{d}^T \text{vec}\left( \underbrace{\mathbf{z}\mathbf{z}^T}_{m\times m}\underbrace{\text{mat}\left(\mathbf{d}\right)}_{m \times t} \underbrace{\left(\mathbf{\Lambda} - h(\mathbf{z}) h(\mathbf{z})^T\right)^T }_{t \times t}\right)
\end{align*}
As highlighted by the shape annotations, this operation can be computed by storing matrices of no larger than $\max(m^2, t^2, mt).$

\newpage
\section{Positive Connections Only}
\label{supp:pos_only_hessian}

For both memory efficiency and conceptual simplicity, instead of learning the Rashomon set over all last layers, we might want to consider one with some parameter tying. In particular, we allow prototypes to connect only with their own class. We begin with notation: let $S_c := \{ i : \psi(i) = c\}$ where $c$ is a class and $\psi(i)$ is a function that returns the class associated with prototype $i$. The constrained last layer forms predictions as:
\begin{align}
    h^{lin}_c(\mathbf{x}) &:= \sum_{i \in S_c } w_i x_i \\
    h(\mathbf{x}) &= \text{softmax}(h^{lin}(\mathbf{x}))
\end{align}

The first partial derivative of cross entropy w.r.t. a parameter $w_i$ is:
\begin{align*}
    \frac{\partial}{\partial w_i} CE(f(\mathbf{x}), \mathbf{y}) &= \frac{\partial}{\partial w_i} \left [ \sum_{k=1}^C y_k \left(h^{lin}_k(\mathbf{x}) - ln(1 + \sum_{k=1}^C \text{exp}(h^{lin}_k(\mathbf{x}))\right)\right]\\
    &= y_{\psi(i)} \frac{\partial}{\partial \omega_i^{(+)}} h^{lin}_{\psi(i)}(\mathbf{x}) - \frac{\text{exp}(h^{lin}_{\psi(i)}(\mathbf{x}))}{1 + \sum_{k=1}^C \text{exp}(h^{lin}_k(\mathbf{x}))} \frac{\partial}{\partial \omega_i^{(+)}} h^{lin}_{\psi(i)}(\mathbf{x})\\
    &= y_{\psi(i)} x_i - \frac{\text{exp}(h^{lin}_{\psi(i)}(\mathbf{x}))}{1 + \sum_{k=1}^C \text{exp}(h^{lin}_k(\mathbf{x}))} x_i\\
    &= (y_{\psi(i)} - h_{\psi(i)}(\mathbf{x})) x_i\\
\end{align*}
For the second derivative, we have:

\begin{align*}
    \frac{\partial}{\partial w_j} \frac{\partial}{\partial w_i} CE(f(\mathbf{x}), \mathbf{y}) &= \frac{\partial}{\partial w_j}(y_{\psi(i)} - h_{\psi(i)}(\mathbf{x})) x_i\\
    &= - x_i \frac{\partial}{\partial w_j} h_{\psi(i)}(\mathbf{x})\\
    &= - x_i \sum_{k=1}^C\frac{\partial h_{\psi(i)}}{\partial h^{lin}_k}\frac{\partial h^{lin}_k}{\partial w_j} \\
    &= - x_i \frac{\partial h_{\psi(i)}}{\partial h^{lin}_{\psi(j)}}\frac{\partial h^{lin}_{\psi(j)}}{\partial w_j} \\
    &= - x_i x_j h_{\psi(i)}(\mathbf{x}) (\delta_{\psi(i)\psi(j)} - h_{\psi(j)}(\mathbf{x}) )\\
\end{align*}

Packaged together, we then have
\begin{align*}
    \mathbf{H} &:= \begin{bmatrix}
        - x^2_1 h_{\psi(1)}(\mathbf{x}) (\delta_{\psi(1)\psi(1)} - h_{\psi(1)}(\mathbf{x}) )  & - x_2 x_1 h_{\psi(1)}(\mathbf{x}) (\delta_{\psi(2)\psi(1)} - h_{\psi(2)}(\mathbf{x}) ) & \hdots\\
        - x_1 x_2 h_{\psi(2)}(\mathbf{x}) (\delta_{\psi(1)\psi(2)} - h_{\psi(1)}(\mathbf{x}) )  & - x^2_2 h_{\psi(2)}(\mathbf{x}) (\delta_{\psi(2)\psi(2)} - h_{\psi(2)}(\mathbf{x}) ) & \hdots\\
        \hdots\\
        - x_1 x_d h_{\psi(d)}(\mathbf{x}) (\delta_{\psi(1)\psi(d)} - h_{\psi(1)}(\mathbf{x}) )  & - x_2 x_d h_{\psi(d)}(\mathbf{x}) (\delta_{\psi(2)\psi(d)} - h_{\psi(2)}(\mathbf{x}) )  & \hdots
    \end{bmatrix}\\
    &= \begin{bmatrix}
        h_{\psi(1)}(\mathbf{x}) (\delta_{\psi(1)\psi(1)} - h_{\psi(1)}(\mathbf{x}) ) &  h_{\psi(1)}(\mathbf{x}) (\delta_{\psi(2)\psi(1)} - h_{\psi(2)}(\mathbf{x}) ) & \hdots\\
         h_{\psi(2)}(\mathbf{x}) (\delta_{\psi(1)\psi(2)} - h_{\psi(1)}(\mathbf{x}) )   & h_{\psi(2)}(\mathbf{x}) (\delta_{\psi(2)\psi(2)} - h_{\psi(2)}(\mathbf{x}) )  & \hdots\\
        \hdots\\
        h_{\psi(d)}(\mathbf{x}) (\delta_{\psi(1)\psi(d)} - h_{\psi(1)}(\mathbf{x}) ) & h_{\psi(d)}(\mathbf{x})(\delta_{\psi(2)\psi(d)} - h_{\psi(2)}(\mathbf{x}) )& \hdots
    \end{bmatrix} \odot -\mathbf{x}\mathbf{x}^T 
\end{align*}
where $\odot$ denotes a Hadamard product.

\clearpage
\section{Sampling From a Rashomon Set After Requiring Prototypes}
Let $J$ denote a set of prototype indices we wish to require, such that we constrain $[\text{coef}(\mathbf{p}_j) \geq \alpha]\forall{j \in J}$ where $\text{coef}(\mathbf{p}_j)$ denotes the last layer coefficient associated with prototype $\mathbf{p}_j$ and $\alpha \in \mathbb{R}$ is the minimum acceptable coefficient. Given this information, there are a number of ways to formulate requiring prototypes as a convex optimization problem. One such form, which we use, is to optimize the following:

\begin{align*}
    \min_{\mathbf{w}_h}\frac{1}{2(\theta - \bar{\ell}(\mathbf{w}_h^*))}(\mathbf{w}_h - \mathbf{w}_h^*)^T\mathbf{H}(\mathbf{w}_h - \mathbf{w}_h^*) \hspace{0.1in}
    \textrm{s.t. } [\mathbf{e}_j^T\mathbf{w}_h \geq \alpha] \forall j \in J
\end{align*}

We then check whether the result for $(\mathbf{w}_h - \mathbf{w}_h^*)^T\mathbf{H}(\mathbf{w}_h - \mathbf{w}_h^*)$ satisfies the constraint of being $< 1$. If so, we've found weights within our Rashomon set approximation. If not, we've proved no such solution exists. 

\label{supp:sampling_halfspace}

\clearpage
\section{Training Details for Reference ProtoPNets}
\label{supp:training_ref_protopnets}
\ourmethod{} is created by first training a reference ProtoPNet, then calculating a subset of the Rashomon set based on that reference model. Each reference ProtoPNet was trained using the Bayesian hyperparameter optimization framework described in \cite{willard2024looks}. We used cosine similarity for prototype comparisons. We trained each backbone using four training phases, as described in \cite{chen2019looks}: warm up, in which only the prototype layer and add-on layers (additional convolutional layers appended to the end of the backbone of a ProtoPNet) are optimized; joint, in which all backbone, prototype layer, and add-on layer parameters are optimized; project, in which prototypes are set to be exactly equal to their nearest neighbors in the latent space; and last-layer only, in which only the final linear layer is optimized.

We trained each model to minimize the following loss term:
\begin{align*}
    \ell_{total} = CE + \lambda_{clst} \ell_{clst} + \lambda_{sep}\ell_{sep} + \lambda_{ortho}\ell_{ortho},
\end{align*}
where each $\lambda$ term is a hyperparameter coefficient, $CE$ is the standard cross entropy loss, $\ell_{clst}$ and $\ell_{sep}$ are the cluster and separation loss terms from \cite{chen2019looks} adapted for cosinse similarity, and $\ell_{ortho}$ is orthogonality loss as defined in \cite{donnelly2022deformable}. In particular, our adaptations of cluster and separation loss are defined as:
\begin{align*}
    \ell_{clst} := \frac{1}{n} \sum_{i=1}^n \max_{j \in \{1, \hdots, m\} :  \text{class}(\mathbf{p}_j) = y_i} g_j(f(\mathbf{X}_i))\\ 
    \ell_{sep} := \frac{1}{n} \sum_{i=1}^n \max_{j \in \{1, \hdots, m\} :\text{class}(\mathbf{p}_j) \neq y_i} g_j(f(\mathbf{X}_i)),
\end{align*}
where $\text{class}(\mathbf{p}_j)$ is a function that returns the class with which prototype $\mathbf{p}_j$ is associated. During last-layer only optimization, we additionally minimize the $\ell_1$ norm of the final linear layer weights with a hyperparameter coefficient $\lambda_{\ell_1}.$
For each of our experiments, we performed Bayesian optimization with the prior distributions specified in Table \ref{tab:main_sweep_priors} and used the model with the highest validation accuracy following projection as our reference ProtoPNet. We deviated from this selection criterion only for the user study; for that experiment, we selected a model with a large gap between train and validation accuracy (indicating overfitting to the induced confounding), and visually confirmed the use of confounded prototypes.

\begin{table}
    \centering
    \begin{tabular}{ p{0.2\linewidth}| p{0.28\linewidth}|p{0.48\linewidth}}
        Hyperparameter Name & Distribution & Description\\
        \hline
        pre\_project\_phase\_len & Integer Uniform; Min=3, Max=15 & The number of warm and joint optimizationm epochs to run before the first prototype projection is performed\\
        post\_project\_phases & Fixed value; 10 & Number of times to perform projection and the subsequent last layer only and joint optimization epochs\\
        phase\_multiplier & Fixed value; 1 & Amount to multiply each number of epochs by, away from a default of 10 epochs per training phase\\
        lr\_multiplier& Normal; Mean=1.0, Std=0.4 & Amount to multiply all learning rates by, relative to the reference values in \cite{willard2024looks}\\
        joint\_lr\_step\_size & Integer Uniform; Min=2, Max=10 & The number of training epochs to complete before each learning rate step, in which each learning rate is multiplied by 0.1\\
        num\_addon\_layers & Integer Uniform; Min=0, Max=2 & The number of additional convolutional layers to add between the backbone and the prototype layer\\
        latent\_dim\_multiplier\_exp & Integer Uniform; Min=-4, Max=1 & If num\_addon\_layers is not 0, dimensionality of the embedding space will be multiplied by $2^{\text{latent\_dim\_multiplier}}$ relative to the original embedding dimension of the backbone \\
        num\_prototypes\_per\_class & Integer Uniform; Min=8, Max=16 & The number of prototypes to assign to each class\\
        cluster\_coef & Normal; Mean=-0.8, Std=0.5 & The value of $\lambda_{clst}$\\
        separation\_coef & Normal; Mean=0.08, Std=0.1 & The value of $\lambda_{sep}$\\
        l1\_coef & Log Uniform; Min=0.00001, Max=0.001 & The value of $\lambda_{\ell_1}$\\
        orthogonality\_coef & Log Uniform; Min=0.00001, Max=0.001 & The value of $\lambda_{ortho}$\\
    \end{tabular}
    \caption{Prior distributions over hypeparameters for the Bayesian sweep used to train all reference ProtoPNets except for the one in the user study.}
    \label{tab:main_sweep_priors}
\end{table}

\clearpage
\section{Hardware Details}
\label{supp:hardware_details}
All of our experiments were run on a large institutional compute cluster. We trained each reference ProtoPNet using a single NVIDIA RTX A5000 GPU with CUDA version 12.4, and ran other experiments using a single NVIDIA RTX A6000 GPU with CUDA version 12.4.

\section{Confounding Details}
\label{supp:confounding_details}
\begin{figure}[b]
    \centering
    \includegraphics[width=1.0\linewidth]{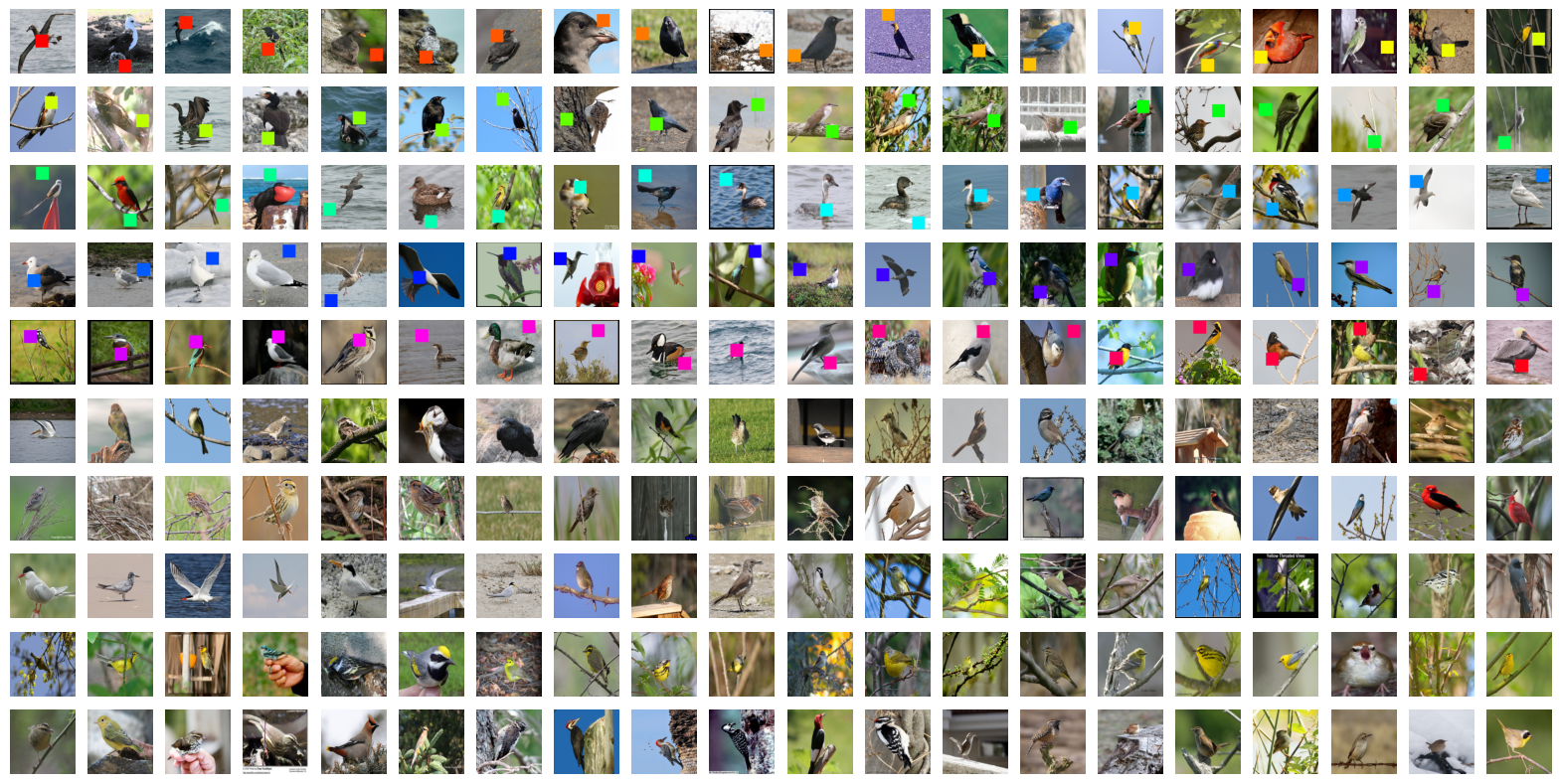}
    \caption{One example image from each class in CUB200, with confounding patches added as in the user study. The first 100 classes receive random confounding patches, and the second 100 are unaltered.}
    \label{fig:confounding_patches_viz}
\end{figure}

To run our user study, we trained a reference ProtoPNet over a version of CUB200 in which a confounding patch has been added to each training image. For each training image coming from one of the first 100 classes, we add a color patch with width and height equal to $1/5$ those of the image to a random location in the image. The color of each patch is determined by the class of the training image; class 0 is set to the initial value fo the HSV color map in matplotlib, class 1 to the color $1/100^{\text{th}}$ further along this color map, and so on. A sample from each class with this confounding is shown in Figure \ref{fig:confounding_patches_viz}. Images from the validation and test splits of the dataset were not altered.

\clearpage
\section{User Study Details}
\label{supp:user_study}
In this section, we describe our user study in further detail. 

\subsection{Setup}
We followed the procedure described in Appendix \ref{supp:confounding_details} to create a confounded version of the CUB200 training set, and fit a ProtoPNet with a VGG-19 backbone over this confounded set. We created this model using a Bayesian hyperparameter sweep, but rather than selecting the model with the highest validation accuracy, we selected the model with the largest gap between its train and validation accuracy, as this indicates overfitting to the confounding patches added to the training set. The hyperparameters used for the selected model are shown in Table \ref{tab:confounded_hyperparams}. Each ProtoPDebug model trained in this experiment used these hyperparameters, and the prescribed ``forbid loss'' from  \cite{bontempelli2023concept} with a coefficient of 100 as in \cite{bontempelli2023concept}.

\begin{figure}
    \centering
    \includegraphics[width=0.87\linewidth]{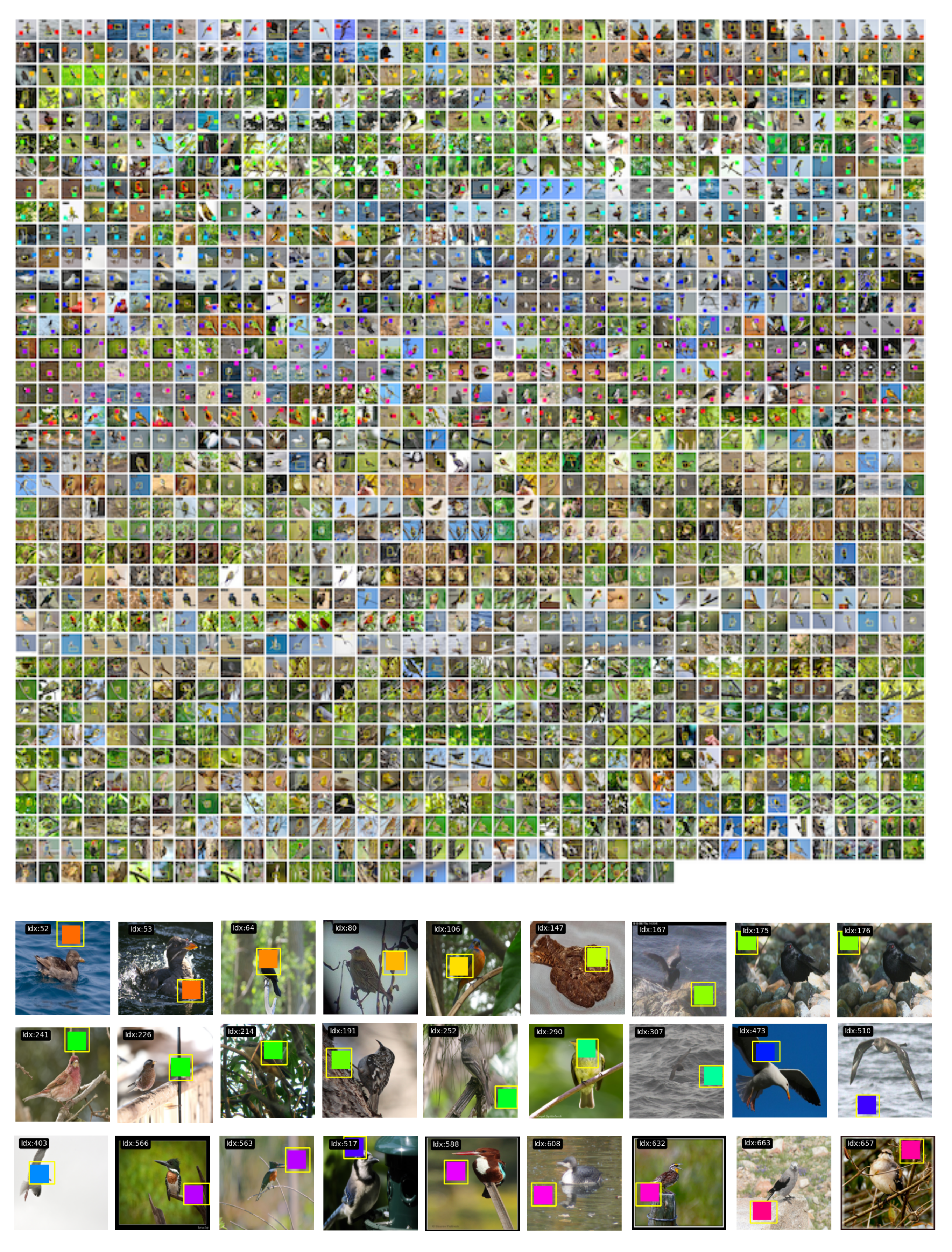}
    \caption{(\textbf{Top}) All $1509$ prototypes used by the confounded model from the user study. (\textbf{Bottom}) The $27$ clearly confounded prototypes we identified as the ``gold standard'' for users to remove.}
    \label{fig:gold_std_removals}
\end{figure}

\begin{figure}
    \centering
    \includegraphics[width=1.0\linewidth]{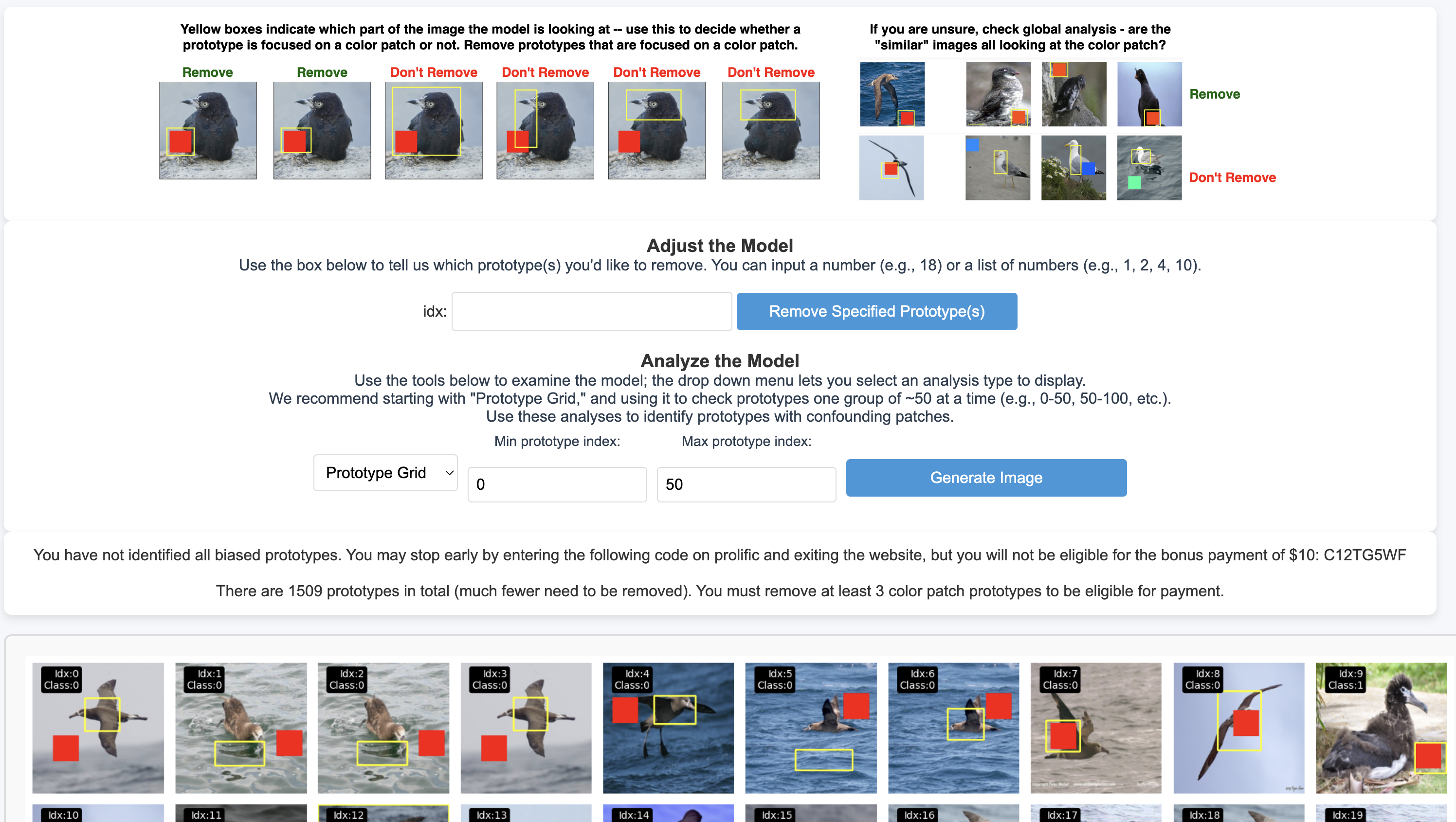}
    \caption{A screenshot of the interface for our user study. Users were tasked with identifying and removing confounded prototypes.}
    \label{fig:ui_screenshot}
\end{figure}

We visually examined the prototypes learned by this model to identify the minimal set of ``gold standard'' prototypes we expected users to remove. We found that, of 1509 prototypes, 27 were clearly confounded, with bounding boxes focused entirely on a confounding patch and global analyses illustrating that the three most similar training patches to each prototype were also confounding patches. Figure \ref{fig:gold_std_removals} shows all prototypes used by this model, as well as the 27 ``gold standard'' confounded prototypes we identified. Figure \ref{fig:ui_screenshot} shows a screenshot of the interface for our user study.

\subsection{Recruitment and Task Description}
Users were recruited from the crowd sourcing platform Prolific, and were tasked with removing prototypes that clearly focused on confounding patches. The complete, anonymized informed consent text shown to users -- which provides instructions -- was as follows:

\begin{displayquote}

This research study is being conducted by \textbf{REDACTED}. This research examines whether a novel tool can be used to remove obvious errors in an AI model. You will be asked to use the tool to remove obvious errors, a task that will take approximately 30 minutes and no longer than an hour. 
Your participation in this research study is voluntary. 
        
You may withdraw at any time, but you will only be paid if you remove at least 10\% of color-patch prototypes and do not remove more than 2 non-color-patch prototypes. 
After completing the survey, you will be paid at a rate of \$15/hour of work through the Prolific platform.
        
In accordance with Prolific policies, we may reject your work if the task was not completed correctly, or the instructions were not followed. 
A bonus payment of \$10 will be offered if all errors are identified and corrected within 30 minutes. 
        
There are no anticipated risks or benefits to participants for participating in this study. 
Your participation is anonymous as we will not collect any information that the researchers could identify you with.
        
If you have any questions about this study, please contact \textbf{REDACTED} and include the term “Prolific Participant Question” in your email subject line. For questions about your rights as a participant contact \textbf{REDACTED}.
        
If you consent to participate in this study please click the “$\rangle \rangle$” below to begin the survey.
\end{displayquote}

A total of 51 Prolific users completed our survey. Of these, 20 submissions were rejected for either 1) failing to identify at least 10\% of the gold standard confounded prototypes (i.e., identified 2 or fewer) or 2) removing more than 2 prototypes drawn from images with no confounding patch. A total of $4$ users qualified for and received the \$10 bonus payment.

\begin{table}
    \centering
    \begin{tabular}{ p{0.2\linewidth}| p{0.2\linewidth}|p{0.53\linewidth}}
        Hyperparameter Name & Value & Description\\
        \hline
        pre\_project\_phase\_len & 11 & The number of warm and joint optimization epochs to run before the first prototype projection is performed\\
        post\_project\_phases & 10 & Number of times to perform projection and the subsequent last layer only and joint optimization epochs\\
        phase\_multiplier & 1 & Amount to multiply each number of epochs by, away from a default of 10 epochs per training phase\\
        lr\_multiplier& 0.89 & Amount to multiply all learning rates by, relative to the reference values in \cite{willard2024looks}\\
        joint\_lr\_step\_size & 8 & The number of training epochs to complete before each learning rate step, in which each learning rate is multiplied by 0.1\\
        num\_addon\_layers & 1 & The number of additional convolutional layers to add between the backbone and the prototype layer\\
        latent\_dim\_multiplier\_exp & -4 & If num\_addon\_layers is not 0, dimensionality of the embedding space will be multiplied by $2^{\text{latent\_dim\_multiplier}}$ relative to the original embedding dimension of the backbone \\
        num\_prototypes\_per\_class & 14 & The number of prototypes to assign to each class\\
        cluster\_coef & -1.2 & The value of $\lambda_{clst}$\\
        separation\_coef & 0.03 & The value of $\lambda_{sep}$\\
        l1\_coef & 0.00001 & The value of $\lambda_{\ell_1}$\\
        orthogonality\_coef & 0.0004 & The value of $\lambda_{ortho}$\\
    \end{tabular}
    \caption{Hyperparameter values used to construct the confounded ProtoPNet for the user study.}
    \label{tab:confounded_hyperparams}
\end{table}

\clearpage
\section{\ourmethod{} Meets Feedback Better than ProtoPDebug}
\label{supp:debug_fails_feedback}
\begin{figure}
    \centering
    \includegraphics[width=0.7\linewidth]{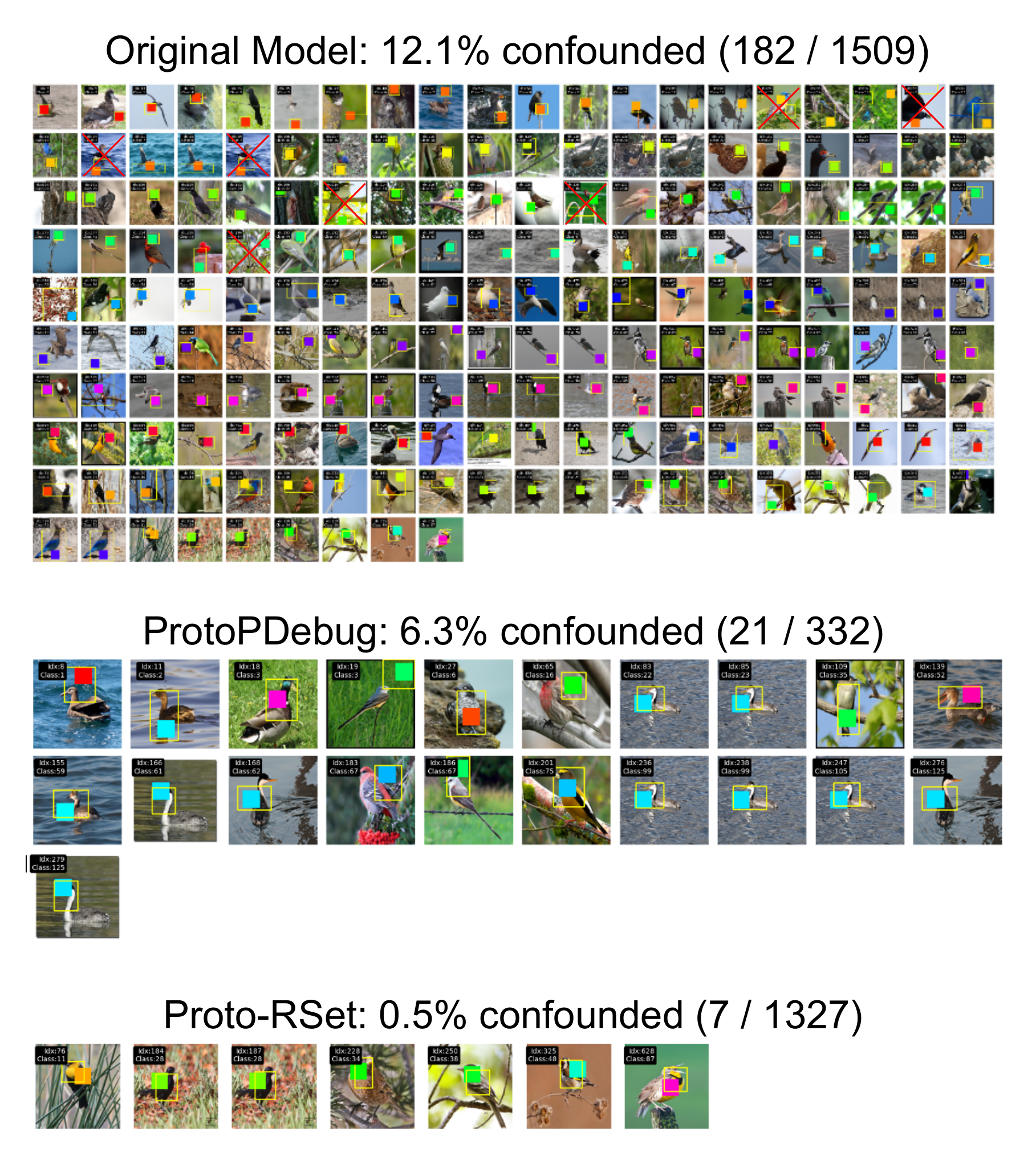}
    \caption{All confounded prototypes from each of a reference ProtoPNet, a model produced by ProtoPDebug, and a model produced by \ourmethod. A prototype is considered confounded if the majority of the bounding box is covered by a confounding patch. \ourmethod{} and ProtoPDebug were shown the feedback obtained from the same random user. A substantially larger proportion of the prototypes from ProtoPDebug are confounded than the prototypes from \ourmethod.}
    \label{fig:protodebug_confounded}
\end{figure}

Here, we use results from the user study to highlight a key advantage in \ourmethod{} over ProtoPDebug: \ourmethod{} guarantees that user constraints are met when possible. We selected a random user who successfully identified all ``gold standard'' confounded prototypes (see Figure \ref{fig:gold_std_removals}). With this user's feedback, we created two models: one using ProtoPDebug and one from \ourmethod{}. We examined all prototypes used by each model. In both models, we identified every prototype for which the majority of the prototype's bounding box focused on a confounding color patch. Figure \ref{fig:protodebug_confounded} shows every confounded prototype from the original reference model, the model produced by ProtoPDebug, and the model produced by \ourmethod. Note that, for the original reference model, we visualize every prototype marked as confounded by the user plus those we identified as confounded. We mark ``false positive'' user feedback (prototypes that were removed, but do not focus on confounding patches) with a red ``X,'' and do not count them toward the total number of confounded prototypes for the original model.

We found that a substantial portion of the prototypes used by ProtoPDebug -- 6.3\% of all prototypes --  still used confounded prototypes despite user feedback. In contrast, only 0.5\% of the prototypes used by \ourmethod{} focused on confounding patches. Moreover, the 7 confounded prototypes that remain after applying \ourmethod{} would not be present if the user had identified them; this is not guaranteed for the 21 confounded prototypes of ProtoPDebug.

\clearpage
\section{Detailed Parameter Values for \ourmethod}
\label{supp:rset_params}

Here, we briefly describe the parameters used when computing the \ourmethod's from each experimental section. Each Rashomon set was defined with respect to an $\ell_2$ regularized cross entropy loss  $\bar{\ell}(\mathbf{w}_h) = CE(\mathbf{w_h}) + \lambda \|\mathbf{w}_h\|_2,$ with $\lambda=0.0001.$ In all of our experiments except for the user study, we used a Rashomon parameter of $\theta=1.1 \bar{\ell}(\mathbf{w}_h^*);$ for the user study, we set $\theta=1.2 \bar{\ell}(\mathbf{w}_h^*)$ to account for the fact that the training set was confounded, meaning training loss was a less accurate indicator of model performance. 
We estimated the empirical loss minimizer $\mathbf{w}_h^*$ using stochastic gradient descent with a learning rate of $1.0$ for a maximum of $5,000$ epochs. We stopped this optimization early if loss decreased by no more than $10^{-7}$ between epochs. 

\clearpage
\section{Evaluating \ourmethod's Ability to Require Prototypes}
\label{supp:requiring_prototypes}
In this section, we evaluate the ability of \ourmethod{} to require prototypes using the method described in the main paper. We match the experimental setup from the main prototype removal experiments; namely, we evaluate \ourmethod{} over three fine-grained image classification datasets (CUB-200 \cite{WahCUB_200_2011}, Stanford Cars \cite{KrauseStarkDengFei-Fei_3DRR2013}, and Stanford Dogs \cite{khosla2011novel}), with ProtoPNets trained on six distinct CNN backbones (VGG-16 and VGG-19 \cite{vgg}, ResNet-34 and ResNet-50 \cite{resnet}, and DenseNet-121 and DenseNet-161 \cite{densenet}) considered in each case. For each dataset-backbone combination, we applied the Bayesian hyperparameter tuning regime of \cite{willard2024looks} for 72 GPU hours and used the best model found in terms of validation accuracy after projection as a reference ProtoPNet. For a full description of how these ProtoPNets were trained, see Appendix \ref{supp:training_ref_protopnets}.

In each of the following experiments, we start with a well-trained ProtoPNet and iteratively require that up to $100$ random prototypes have coefficient greater than $\tau$, where $\tau$ is the mean value of the non-zero entries in the reference ProtoPNet's final linear layer. If we find that no protoype can be required from the model while remaining in the Rashomon set, we stop this procedure early. This occurred $n$ times.

We consider two baselines for comparison in each of the following experiments: naive prototype requirement, in which the correct-class last-layer coefficient for each required prototype is set to $\tau$ and no further adjustments are made to the model, and naive prototype requirement with retraining, in which a similar requirement procedure is applied, but the last-layer of the ProtoPNet is retrained (with all other parameters held constant) after prototype requirement. In the second baseline, we apply an $\ell_1$ \textit{reward} to coefficients corresponding to required prototypes to prevent the model from forgetting them and train the last layer until convergence, or for up to 5,000 epochs -- whichever comes first. By $\ell_1$ reward, we mean that this quantity is \textit{subtracted} from the overall loss value, so that a larger value for these entries decreases loss.

\begin{figure}[b]
    \centering
    \includegraphics[width=1.0\linewidth]{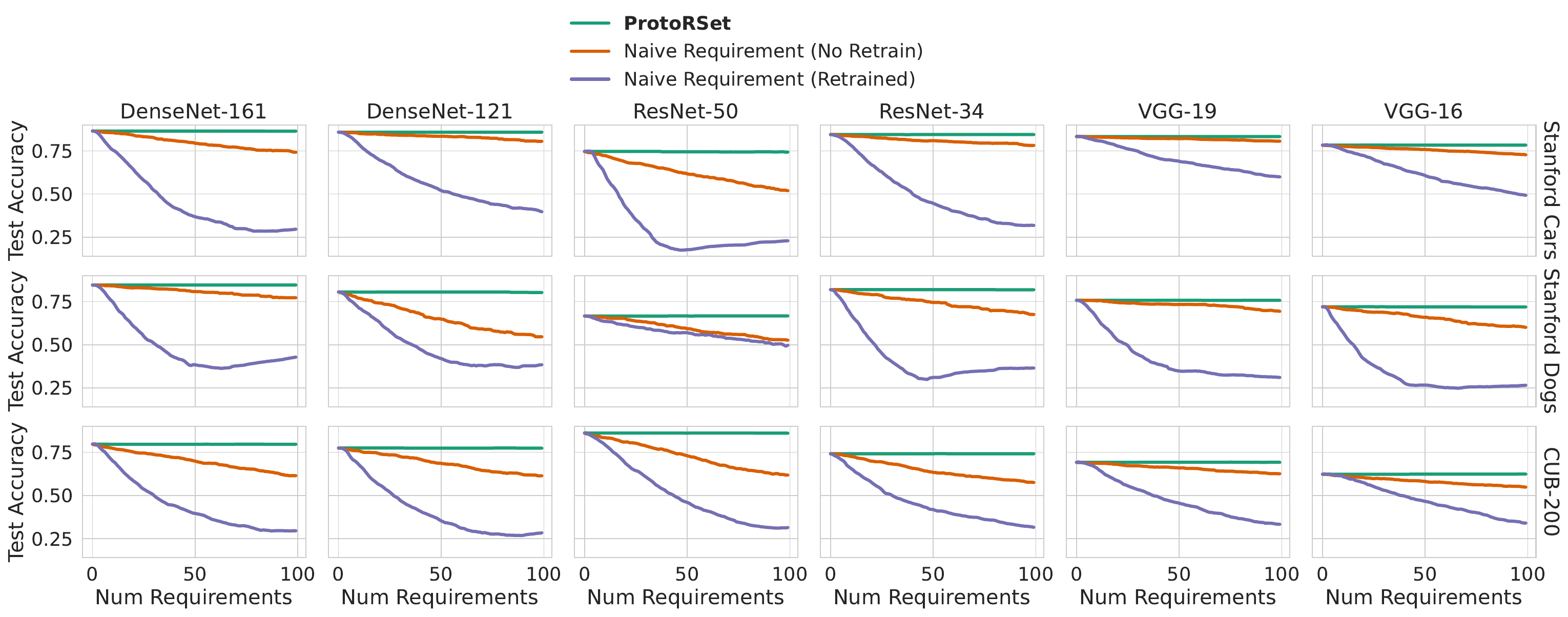}
    \caption{Change in test accuracy as random prototypes are required. In all cases, we see that requiring prototypes using ProtoRSet maintains or slightly improves the accuracy of the original model. Naively requiring prototypes either with or without retraining, on the other hand, dramatically reduces model performance.}
    \label{fig:requirement_accuracy_fig}
\end{figure}

\textbf{\ourmethod{} Produces Accurate Models.} Figure \ref{fig:requirement_accuracy_fig} presents the test accuracy of each model produced in this experiment as a function of the number of prototypes required. We find that, across all six backbones and all three datasets, \textbf{\ourmethod{} maintains test accuracy as constraints are added.} This stands in stark contrast to direct requirement of undesired prototypes, which dramatically reduces performance in all cases. The test accuracy of models produced by \ourmethod{} is maintained \textbf{in all cases.}

\begin{figure*}[t]
    \centering
    \includegraphics[width=1.0\linewidth]{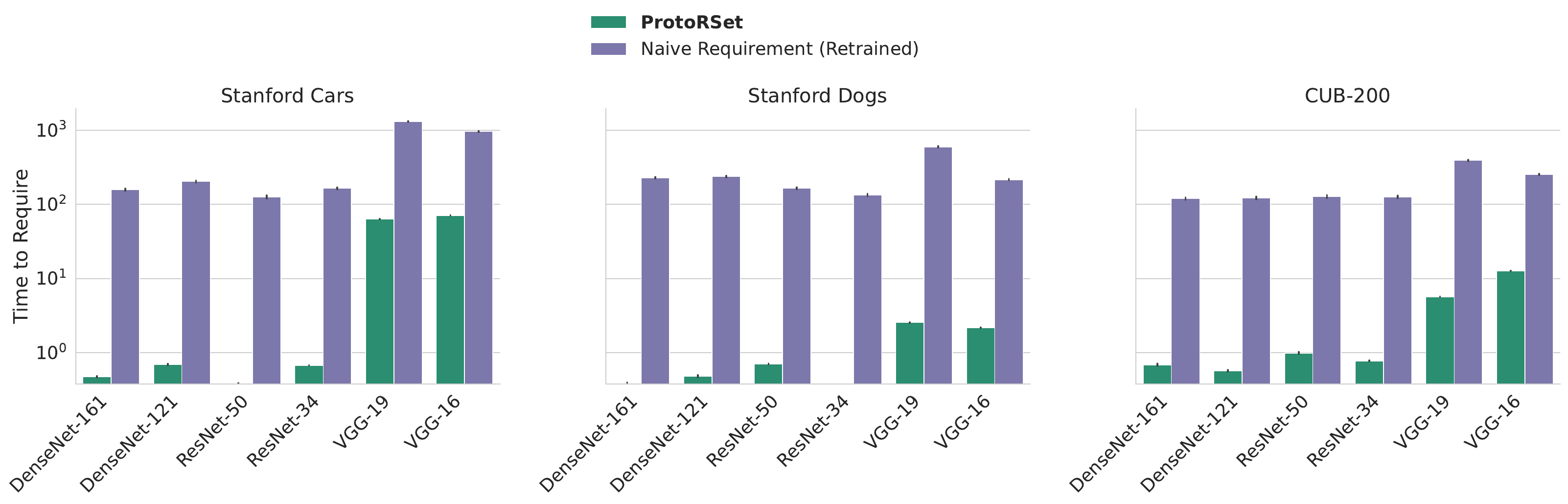}
    \caption{Time in seconds to produce a model meeting requirement constraints, averaged over 100 iterations of requirement. In all cases, ProtoRSet meets the prototype requirement constraint faster than the naive method (requiring a prototype then retraining the last layer). We exclude naive requirement without retraining from the chart because it is simply updating a value in an array, and as such is nearly instantaneous.}
    \label{fig:requirement_timing_fig}
\end{figure*}

\textbf{\ourmethod{} is Fast.} Figure \ref{fig:requirement_timing_fig} presents the time necessary to require a prototype using \ourmethod{} versus naive prototype requirement with retraining. We observe that, across all backbones and datasets, \ourmethod{} requires prototypes orders of magnitude faster than retraining a model. In contrast to prototype removal, we observe that the time necessary to require prototypes is heavily dependent on the reference ProtoPNet. Naive requirement without retraining tends to be faster, but at the cost of substantial decreases in accuracy.

\clearpage
\section{Evaluating the Sampling of Additional Prototypes}
\label{supp:sampling_prototypes}

In this section, we evaluate whether the prototype sampling mechanism described in subsection \ref{subsec:sampling_prototypes} allows users to impose additional constraints by revisiting the skin cancer classification case study from subsection \ref{subsec:skin_cancer}. In subsection \ref{subsec:skin_cancer}, we attempted to remove 10 of the original 21 prototypes used by the model, and found that we were unable to remove the one of these prototypes without sacrificing accuracy. 

\begin{figure}[b]
    \centering
    \includegraphics[width=1.0\linewidth]{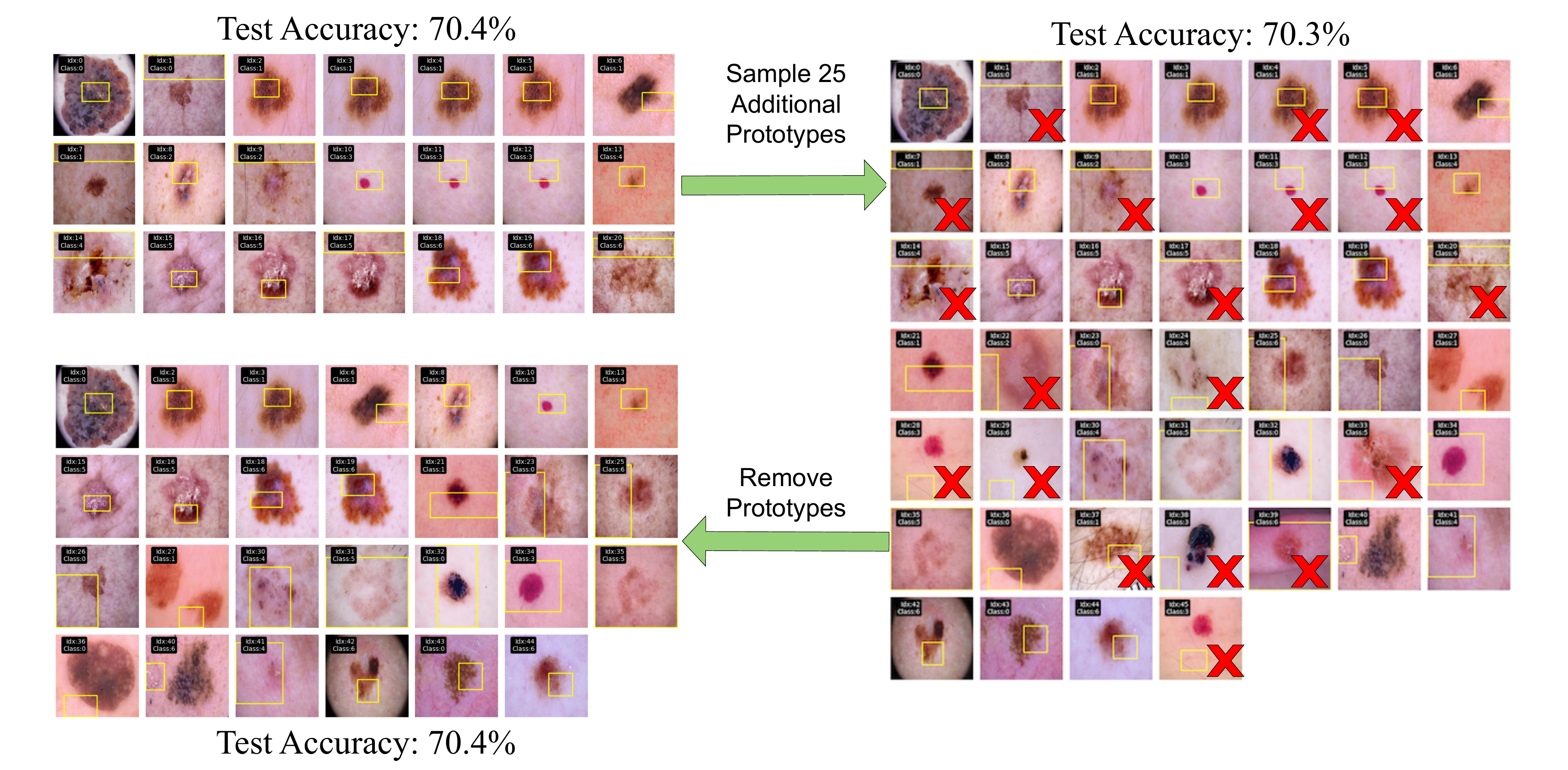}
    \caption{The process followed to remove all desired prototypes from a model for skin cancer classification. Starting from a reference ProtoPNet with 21 prototypes (Top Left), we first sampled 25 additional prototypes to produce a set of 46 candidate prototypes (Right). We removed 19 prototypes from this set of candidates, each of which is annotated with a red ``X''. Removing these prototypes using \ourmethod{} resulted in a model with 27 non-confounded prototypes that matched the test accuracy of the original model.}
    \label{fig:ham_sampling}
\end{figure}

We sampled 25 additional prototypes using the mechanism described in subsection \ref{subsec:sampling_prototypes} using the same reference ProtoPNet, and analyzed the resulting model for any additional background or duplicate prototypes. We found that 9 out of the 25 additional prototypes focused primarily on the background. We removed all 10 of the original target prototypes and these 9 new ones, resulting in a model with 27 prototypes in total, as shown in Figure \ref{fig:ham_sampling}. \textbf{By sampling additional prototypes, \ourmethod{} was able to meet user constraints that were not possible given the original set of prototypes.} This model achieved identical test accuracy to the original model. Recalling that accuracy dropped substantially when removing the 10 target prototypes without sampling alternatives, this demonstrates that sampling more prototypes increases the flexibility of  \ourmethod.

\clearpage
\section{Additional Visualizations of Model Refinement}
\label{supp:refinement visualizations}
In this section, we provide additional visualizations of model reasoning before and after user constraints are added. We consider the ResNet-50 based ProtoPNet trained for CUB-200 that we used for our experiments in the main body. The changes in prototype class-connection weight show that substantial changes to model reasoning are achieved when editing models using \ourmethod{}.

\begin{figure}[b]
    \centering
    \includegraphics[width=1.0\linewidth]{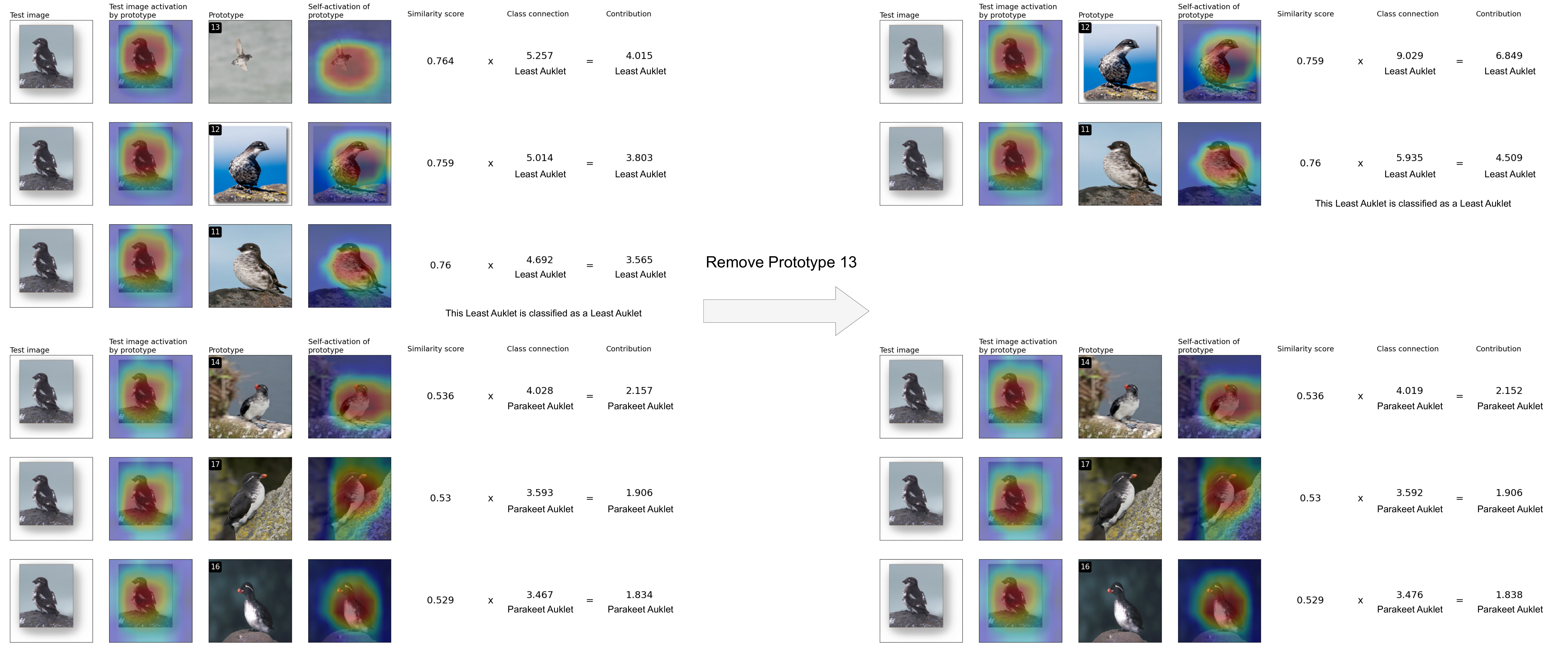}
    \caption{Reasoning process for a ProtoPNet classifying a Least Auklet before (Left) and after (Right) prototype 13 is removed using \ourmethod. In each column, we present the reasoning for the predicted class (Top) and the class with the second highest logit (Bottom).}
    \label{fig:removal_example}
\end{figure}

Figure \ref{fig:removal_example} presents the reasoning process of the model in classifying a Least Auklet before and after an arbitrary prototype is removed. In this case, we see that Proto-RSet substantially adjusted the weight assigned to prototypes of the same class as the removed prototype, and slightly adjusted the coefficient on prototypes from the class with the second highest logit -- in this case, the Parakeet Auklet class. Notably, the weight assigned to prototype 12 increases from 5.014 to 9.029.

\begin{figure}[b]
    \centering
    \includegraphics[width=1.0\linewidth]{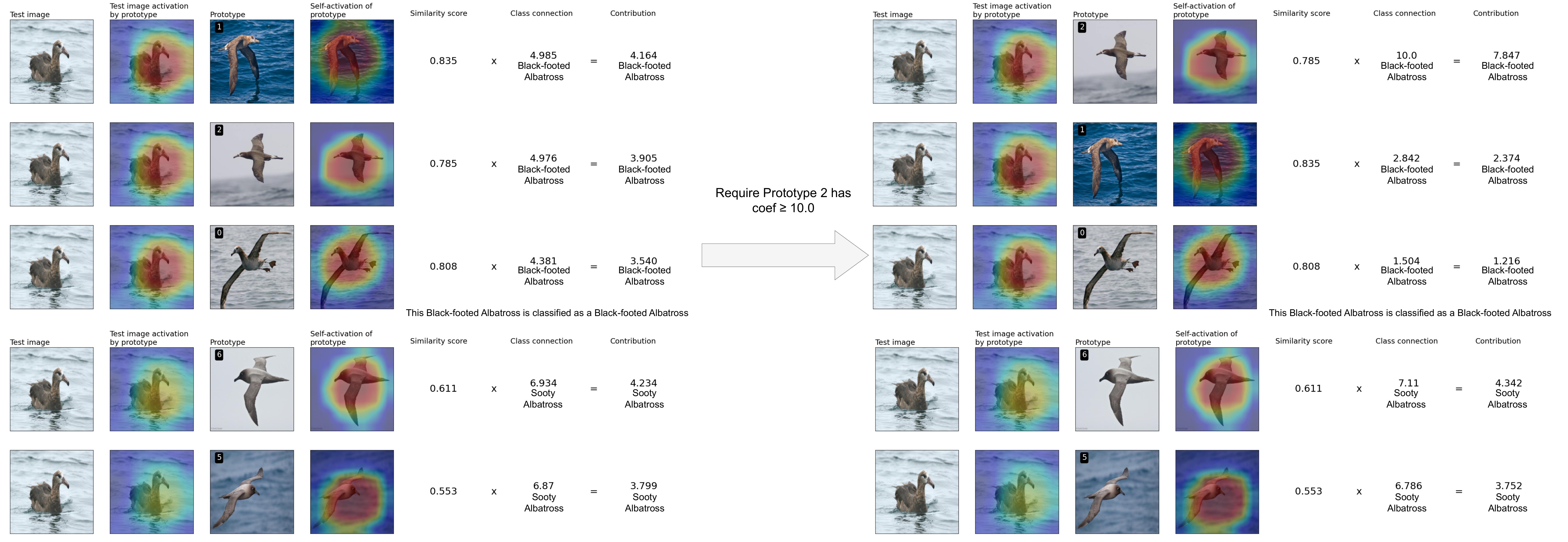}
    \caption{Reasoning process for a ProtoPNet classifying a Black-footed Albatross before (Left) and after (Right) the model is constrained such that prototype 2 has a coefficient of at least 10 using \ourmethod. In each column, we present the reasoning for the predicted class (Top) and the class with the second highest logit (Bottom).}
    \label{fig:requirement_example}
\end{figure}

Figure \ref{fig:requirement_example} presents the reasoning process of the model in classifying a Black-footed Albatross before and after we require a large coefficient be assigned to an arbitrary prototype. We see that Proto-RSet substantially decreased the weight assigned to all other prototypes of the same class as the upweighted prototype, and moderately changed the coefficient on prototypes from the class with the second highest logit -- in this case, the Sooty Albatross class. In particular, the weight assigned to prototype 6 was increased from 6.934 to 7.11, and the weight on prototype 5 was decreased from 6.87 to 6.786.

\end{document}